\documentclass[10pt,letterpaper]{article}

\usepackage[letterpaper,textwidth=6in,textheight=9in]{geometry}
\usepackage{times}    
\usepackage[round,authoryear]{natbib}
\usepackage{titling}
\usepackage{amsmath,amssymb}
\usepackage{wrapfig}
\usepackage{booktabs,tabularx}
\usepackage{threeparttable}
\usepackage{multirow}
\usepackage{placeins}
\usepackage{caption}
\usepackage{pgfplots}
\usepackage{hyperref}

\usepackage[T1]{fontenc}

\newcommand{\copyplotfont}{%
  \fontencoding{T1}\fontfamily{DejaVuSans-TLF}\selectfont
}

\pretitle{\begingroup\centering\Large\scshape}
\posttitle{\par\endgroup\vskip 0.5em}

\usepackage{amsmath,amsfonts,bm}

\def\eqref#1{equation~\ref{#1}}

\def\1{\bm{1}}

\DeclareMathAlphabet{\mathsfit}{\encodingdefault}{\sfdefault}{m}{sl}
\SetMathAlphabet{\mathsfit}{bold}{\encodingdefault}{\sfdefault}{bx}{n}

\pgfplotsset{compat=1.18}
\usetikzlibrary{plotmarks,arrows.meta}
\pgfdeclareplotmark{synthetic star}{
  \pgfpathmoveto{\pgfpointpolar{90}{\pgfplotmarksize}}
  \pgfpathlineto{\pgfpointpolar{126}{.381966\pgfplotmarksize}}
  \pgfpathlineto{\pgfpointpolar{162}{\pgfplotmarksize}}
  \pgfpathlineto{\pgfpointpolar{198}{.381966\pgfplotmarksize}}
  \pgfpathlineto{\pgfpointpolar{234}{\pgfplotmarksize}}
  \pgfpathlineto{\pgfpointpolar{270}{.381966\pgfplotmarksize}}
  \pgfpathlineto{\pgfpointpolar{306}{\pgfplotmarksize}}
  \pgfpathlineto{\pgfpointpolar{342}{.381966\pgfplotmarksize}}
  \pgfpathlineto{\pgfpointpolar{378}{\pgfplotmarksize}}
  \pgfpathlineto{\pgfpointpolar{414}{.381966\pgfplotmarksize}}
  \pgfpathclose\pgfusepathqfillstroke}
\definecolor{c13266f}{RGB}{19,38,111}
\definecolor{c17becf}{RGB}{23,190,207}
\definecolor{c1f77b4}{RGB}{31,119,180}
\definecolor{c2072b1}{RGB}{32,114,177}
\definecolor{c2878a5}{RGB}{40,120,165}
\definecolor{c2a78d6}{RGB}{42,120,214}
\definecolor{c2ca02c}{RGB}{44,160,44}
\definecolor{c333333}{RGB}{51,51,51}
\definecolor{c4c4c4c}{RGB}{76,76,76}
\definecolor{c595959}{RGB}{89,89,89}
\definecolor{c59bfc0}{RGB}{89,191,192}
\definecolor{c737373}{RGB}{115,115,115}
\definecolor{c7f7f7f}{RGB}{127,127,127}
\definecolor{c8c564b}{RGB}{140,86,75}
\definecolor{c8c8c8c}{RGB}{140,140,140}
\definecolor{c9467bd}{RGB}{148,103,189}
\definecolor{cbcbd22}{RGB}{188,189,34}
\definecolor{cbfbfbf}{RGB}{191,191,191}
\definecolor{cd62728}{RGB}{214,39,40}
\definecolor{ce377c2}{RGB}{227,119,194}
\definecolor{ce6862f}{RGB}{230,134,47}
\definecolor{ceb6834}{RGB}{235,104,52}
\definecolor{cff7f0e}{RGB}{255,127,14}
\definecolor{cffffff}{RGB}{255,255,255}
\definecolor{c111111}{RGB}{17,17,17}
\definecolor{c13266f}{RGB}{19,38,111}
\definecolor{c17becf}{RGB}{23,190,207}
\definecolor{c1f77b4}{RGB}{31,119,180}
\definecolor{c2072b1}{RGB}{32,114,177}
\definecolor{c2878a5}{RGB}{40,120,165}
\definecolor{c2a78d6}{RGB}{42,120,214}
\definecolor{c2ca02c}{RGB}{44,160,44}
\definecolor{c333333}{RGB}{51,51,51}
\definecolor{c42a69a}{RGB}{66,166,154}
\definecolor{c4c4c4c}{RGB}{76,76,76}
\definecolor{c595959}{RGB}{89,89,89}
\definecolor{c59bfc0}{RGB}{89,191,192}
\definecolor{c737373}{RGB}{115,115,115}
\definecolor{c7f7f7f}{RGB}{127,127,127}
\definecolor{c8c564b}{RGB}{140,86,75}
\definecolor{c8c6d31}{RGB}{140,109,49}
\definecolor{c8c8c8c}{RGB}{140,140,140}
\definecolor{c8e6bb0}{RGB}{142,107,176}
\definecolor{c9467bd}{RGB}{148,103,189}
\definecolor{cbcbd22}{RGB}{188,189,34}
\definecolor{cbfbfbf}{RGB}{191,191,191}
\definecolor{cd62728}{RGB}{214,39,40}
\definecolor{cd94f45}{RGB}{217,79,69}
\definecolor{cd9d9d9}{RGB}{217,217,217}
\definecolor{ce377c2}{RGB}{227,119,194}
\definecolor{ce6862f}{RGB}{230,134,47}
\definecolor{ceb6834}{RGB}{235,104,52}
\definecolor{cff7f0e}{RGB}{255,127,14}
\definecolor{cffffff}{RGB}{255,255,255}

\pgfplotsset{synthetic axis/.style={
  enlargelimits=false,scaled ticks=false,minor tick num=0,
  tick align=outside,axis line style={black!65},
  grid=major,grid style={black!12},
  clip=true,unbounded coords=jump,
  label style={font=\small},tick label style={font=\footnotesize},
  legend style={font=\footnotesize},title style={font=\small},
}}

\usepackage{fix-cm}
\usepackage{pgfplots}
\usetikzlibrary{patterns,plotmarks}

\newcommand{\LMkey}[4]{\tikz[baseline=-.6ex]{\def\LMsty{#2}\def\LMmarks{only marks}\ifx\LMsty\LMmarks\else\draw[color=#1,#2,line width=.7pt] (0,0) -- (.24,0);\fi\draw[color=#1] plot[mark=#3,mark size=1.7pt,mark options={fill=#1}] coordinates {(.12,0)};}\,#4}
\newcommand{\LMhatch}[3]{\tikz[baseline=-.6ex]{\fill[gray] (0,-.04) rectangle (.2,.08);\def\LMpat{#1}\ifx\LMpat\empty\else\fill[pattern=#1,pattern color=white] (0,-.04) rectangle (.2,.08);\fi\draw[gray] plot[mark=#2,mark size=1.7pt] coordinates {(.32,.02)};}\,#3}
\definecolor{lm000000}{HTML}{000000}
\definecolor{lm17BECF}{HTML}{17BECF}
\definecolor{lm1A1A1A}{HTML}{1A1A1A}
\definecolor{lm1F77B4}{HTML}{1F77B4}
\definecolor{lm2B2B2B}{HTML}{2B2B2B}
\definecolor{lm2CA02C}{HTML}{2CA02C}
\definecolor{lm39A99C}{HTML}{39A99C}
\definecolor{lm4CB3F5}{HTML}{4CB3F5}
\definecolor{lm555555}{HTML}{555555}
\definecolor{lm595959}{HTML}{595959}
\definecolor{lm696969}{HTML}{696969}
\definecolor{lm7B3FB0}{HTML}{7B3FB0}
\definecolor{lm7F7F7F}{HTML}{7F7F7F}
\definecolor{lm808080}{HTML}{808080}
\definecolor{lm8C3B00}{HTML}{8C3B00}
\definecolor{lm8C564B}{HTML}{8C564B}
\definecolor{lm92702D}{HTML}{92702D}
\definecolor{lm9467BD}{HTML}{9467BD}
\definecolor{lmBCBD22}{HTML}{BCBD22}
\definecolor{lmD62728}{HTML}{D62728}
\definecolor{lmE377C2}{HTML}{E377C2}
\definecolor{lmE3E3E3}{HTML}{E3E3E3}
\definecolor{lmE6550D}{HTML}{E6550D}
\definecolor{lmEF4B43}{HTML}{EF4B43}
\definecolor{lmFF7F0E}{HTML}{FF7F0E}
\definecolor{lmFFFFFF}{HTML}{FFFFFF}

\usepackage{float}

\makeatletter
\renewcommand\section{\@startsection{section}{1}{\z@}
  {-4.0ex plus -1ex minus -.2ex}
  {2.6ex plus .2ex}
  {\normalfont\large\scshape\raggedright}}

\renewcommand\subsection{\@startsection{subsection}{2}{\z@}
  {-3.7ex plus -1ex minus -.2ex}
  {1.8ex plus .2ex}
  {\normalfont\normalsize\scshape\raggedright}}
\makeatother

\title{Eliciting the Recall Capacity of Linear Attention}
\author{
  Johan Lindqvist$^{1,2,3,4}$\thanks{\texttt{johan.lindqvist@tuebingen.mpg.de}}
  \qquad
  Antonio Orvieto$^{1,2,3}$ \\[0.5em]
  \small $^1$ELLIS Institute Tübingen \quad
  $^2$MPI-IS \quad
  $^3$Tübingen AI Center\quad
  $^4$KTH Royal Institute of Technology \\
  \small Tübingen, Germany
}
\date{}

\begin{document}

\maketitle

\begin{abstract}
\noindent
Linear attention models exhibit a disproportionate deficit relative to softmax attention in associative recall (AR) in language modeling. We investigate this gap through natural language evaluations, synthetic recall tasks, and targeted training interventions. Using a synthetic setting, we quantify the role of state-size in how many items can be recalled from memory and find a similar empirical scaling across the tested linear-attention architectures. Testing this relationship reveals that even a single layer at scale can learn to retrieve thousands of associations across long distances. Yet, standard language model training does not reliably teach a linear model to copy from context, whereas a comparable softmax model learns to do so. Considering this, we strengthen recall supervision during training through in-context sentence repetition and context-sensitive loss reweighting. Both methods improve recall-intensive benchmark performance and narrow the long-distance AR gap to the softmax baseline without changing the architecture, token budget, or number of optimization steps, at a slight cost to general language modeling. These targeted changes to training can thus elicit substantial but underutilized retrieval capacity in linear-attention models. These findings suggest that, although linear-attention architectures have substantial capacity for recall, their inductive biases and the demands of language model training make it difficult to develop this capability without dedicated incentives.

\end{abstract}

\section{Introduction}
Linear attention offers an efficient alternative to softmax attention by compressing the growing key--value cache into a fixed-size recurrent state \citep{pmlr-v119-katharopoulos20a}. This allows inference costs per token to remain constant with sequence length, making these architectures attractive for processing long contexts. However, their ability to use information from earlier in the sequence remains a persistent concern. Even when overall language modeling performance is comparable, linear attention models often underperform on associative recall (AR), where previously encountered information must be retrieved from a partial cue \citep{jelassi2024repeatmetransformersbetter,waleffe2024empiricalstudymambabasedlanguage,arora2024zoology}. This recall gap remains a major obstacle to the wider adoption of fully recurrent architectures and has motivated both changes to recurrent state updates \citep{delta_net,hatamizadeh2026gateddeltanet2decouplingerase} and hybrid models that combine recurrent layers with softmax attention \citep{waleffe2024empiricalstudymambabasedlanguage,merrill2026olmohybrid}.

We begin by investigating recall in language models, comparing Gated Linear Attention \citep{yang2024gatedlinearattentiontransformers}, DeltaNet \citep{delta_net}, Gated DeltaNet \citep{yang2025gateddeltanetworksimproving}, and Kimi Delta Attention \citep{kimiteam2025kimilinearexpressiveefficient} with softmax attention within a shared Transformer-style block. Following \citet{arora2024zoology}, the evaluation measures predictions that complete rare bigrams previously observed in the same context. The deficit is particularly pronounced for these prefix-matching predictions at long retrieval distances, while differences are much smaller when a repeated rare token appears after a different prefix. Similarly, the average gap on the original QA prompts is small, but replacing them with the exact prefix preceding the answer exposes a larger advantage for softmax attention. Together, these results identify prefix-matching retrieval as a particular weakness of the linear-attention models.

What explains this weakness? The fixed-size recurrent state necessarily limits how much information a model can retain, and prior work has linked this constraint to deficits in copying and associative recall \citep{jelassi2024repeatmetransformersbetter,arora2025simplelinearattentionlanguage}. However, a storage limit alone does not establish that capacity is the binding constraint at practical model sizes and context lengths. Beyond storage constraints, prior work has identified architectural inductive biases that can make retrieval harder to learn, including smoother attention distributions that reduce selectivity \citep{zhang2024hedgehogporcupineexpressive} and recency bias that favors recent information over distant context \citep{wang2025understandingmitigatingbottlenecksstate}. This raises two related questions: how much recall capacity do linear-attention models have at language model scale, and how effectively does standard language model training use it?

To investigate the first question, synthetic multi-query associative recall (MQAR) \citep{arora2024zoology} provides a controlled setting for studying retrieval. Across the evaluated linear-attention variants, the number of correctly retrieved associations grows approximately linearly with recurrent state size, while varying state layout or the parameter count at fixed state size has comparatively little effect.
This scaling predicts that a single layer at language-model scale can retrieve thousands of associations from a 16k-token context. We verify this by training models directly on the task.

Copying exhibits similar state-size scaling. In pretrained language models, however, standard training produces different outcomes. The softmax model learns to copy from context in a teacher-forced task, whereas a representative linear-attention model does not do so reliably. Copy fine-tuning nevertheless enables the linear model to reproduce long passages accurately. Together, these results show that linear-attention models can learn substantial recall under direct task supervision, while standard language model training does not reliably develop the same capability.

These findings motivate a return to language model training. Qualifying AR-hits account for only about 1.7\% of tokens in the held-out FineWeb-Edu sample, suggesting that standard pretraining may provide insufficient incentive to learn robust recall. We test this hypothesis by strengthening the retrieval signal through in-context sentence repetition and Thresholded LongCE (TLCE), which upweights predictions that benefit from longer context. Applied during the final 20\% of training, these interventions improve recall with the architecture, state size, token budget, and optimization-step budget unchanged. Their combination closes the long-distance AR-hit loss gap to standard-trained softmax, at a slight cost to general language modeling. Applying both throughout KDA training further improves recall without an additional language modeling penalty. These results show that targeted training can bring substantial existing retrieval capacity into use.

\section{Linear Attention}
Attention \citep{vaswani2023attentionneed} refers to the operation of selectively retrieving information in a sequence by constructing query, key and value tensors from the input. The most commonly used variant is softmax attention, which uses the softmax function as a similarity kernel to compute the output at time-step $t$ as a normalized sum over all preceding positions. Equivalently in simplified\footnote{Leaving out the causal mask and temperature} parallel and sequential forms: 
{\setlength{\abovedisplayskip}{4pt}
 \setlength{\belowdisplayskip}{4pt}
 \begin{align}
    &O = \operatorname{softmax}(QK^\top)V
    \quad\equiv\quad
    &o_t = \sum_{i=0}^{t}\frac{\exp\!\bigl(q_t^\top k_i\bigr)}
               {\sum_{j=0}^{t} \exp\!\bigl(q_t^\top k_j\bigr)} v_i \\
    &Q, K, V, O \in \mathbb{R}^{T,d}
    &q_t, k_t, v_t, o_t \in \mathbb{R}^{d} \notag
    \label{eq:softmax-attn}
 \end{align}
}
In training, computing interactions between all positions scales quadratically with sequence length. During inference, the entire past key--value cache \citep{pope2022efficientlyscalingtransformerinference} has to be retained and attended to by the new query, leading to linear scaling in both compute and memory usage. In an effort to improve scaling to process longer sequences, linear attention \citep{pmlr-v119-katharopoulos20a} removes the softmax function and uses associativity to reformulate the operation\footnote{Simplified to remove normalization, feature maps, and the causal mask} as:
\begin{align}
 &O = (QK^\top)V = Q(K^\top V)
\end{align}
This removes the pairwise interactions and allows the sequential form to be expressed as a recurrence:
\begin{align}
S_t &= \sum_{i=0}^{t} k_i v_i^\top
     = S_{t-1} + k_t v_t^\top, 
o_t = S_t^\top q_t
     = \sum_{i=0}^{t} (q_t^\top k_i)v_i \\
 &q_t,k_t,v_t,o_t \in \mathbb{R}^d,\;
S_t \in \mathbb{R}^{d \times d}. \notag
\end{align}

This form compresses the KV-cache to a fixed-size recurrent state, making per-token inference cost independent of context length. In the recurrent form, linear attention closely resembles classical recurrent neural networks \citep{elman1990finding}, while the associativity of the update permits efficient parallel computation. Building on this core by drawing inspiration from RNNs, subsequent architectures have added components which modify the state between steps while preserving the efficient parallelism.

Two prominent directions are the Forget Gate \citep{yang2024gatedlinearattentiontransformers} and the Delta Rule \citep{schlag2021lineartransformerssecretlyfast, delta_net}. 
\[
\begin{aligned}
S_t &= \alpha_t S_{t-1} + k_t v_t^\top
&& \text{(Forget Gate)} 
\quad\quad S_t &= (I-\beta_tk_t k_t^\top)S_{t-1} + \beta_tk_t v_t^\top
&& \text{(Delta Rule)}
\end{aligned}
\]

Both operators allow the model to erase information from past context, introducing a recency bias and preventing a linear growth of magnitude across sequence length. The forget gate introduces exponential decay. The decay is controlled by the $\alpha_t$ parameter which depends on the input \citep{yang2024gatedlinearattentiontransformers} allowing the model to determine how much of the past to forget based on its current input. The delta rule instead updates the prediction associated with the current key, selectively modifying overlapping stored information.

Related state-space and recurrent architectures \citep{mamba,mamba2,lahoti2026mamba3improvedsequencemodeling,peng2023rwkv,peng2025rwkv7} use similar state updates but different surrounding blocks. We focus on sequence mixers evaluated within a matched Transformer-style block; \autoref{app:sequence_mixer_details} summarizes related update rules and details of layer structure. We survey sequence mixers that use one or both of these two update mechanisms, in particular: Gated Linear Attention \citep{yang2024gatedlinearattentiontransformers}, DeltaNet \citep{delta_net}, Gated DeltaNet \citep{yang2025gateddeltanetworksimproving} and Kimi Delta Attention \citep{kimiteam2025kimilinearexpressiveefficient}. 

\section{Associative Recall}

Associative recall (AR) \citep{hopfield1982neural} is the ability to retrieve information encountered earlier in context using a related partial cue. In language, this supports connecting references to previously introduced entities and recalling facts needed to interpret or continue a passage. For example, after reading ``Maya left her notebook in the kitchen before heading to work,'' correctly answering ``Where did Maya leave her notebook?'' with ``In the kitchen'' constitutes a successful associative recall query.

Empirically, fixed-state architectures are known to underperform on AR compared to softmax attention \citep{jelassi2024repeatmetransformersbetter,waleffe2024empiricalstudymambabasedlanguage,arora2024zoology}, particularly over long distances. This gap is especially consequential because long-context tasks are where their favorable sequence-length scaling promises the greatest benefit, but realizing that benefit requires effectively using information from earlier in the sequence. To investigate the issue using modern sequence mixers, we train a series of LLMs at a standard academic scale of $\sim$340M parameters on 15B tokens of FineWeb-Edu \citep{penedo2024finewebdatasetsdecantingweb}, using the Llama-2 tokenizer \citep{touvron2023llama2openfoundation} and a context length of 4,096 tokens.\footnote{\autoref{app:llm_training_details} provides further details on training and model setup.}
\begin{table}[htbp]
  \centering
  \small
  \setlength{\tabcolsep}{3pt}
  \renewcommand{\arraystretch}{0.85}
  \setlength{\tabcolsep}{3pt}
  \begin{tabular}{@{}lrrrrrr@{}}
    \toprule
    & & & \multicolumn{4}{c}{Loss difference vs.\ softmax (nats)} \\
    \cmidrule(lr){4-7}
    Group & Count & Softmax (nats) & DN & GLA & GDN & KDA \\
    \midrule
    Standard tokens & 15,712,918 & $2.39$ & $+0.04$ & $+0.02$ & $+0.01$ & $-0.02$ \\
    \midrule
    AR-hits    & 273,070    & $0.89$ & $+0.18$ & $+0.22$ & $+0.14$ & $+0.12$ \\
    \midrule
    \quad 1--511 tokens back  & 226,922 & $0.77$ & $+0.12$ & $+0.15$ & $+0.09$ & $+0.07$ \\
    \quad 512+ tokens back    & 46,148  & $1.45$ & $+0.46$ & $+0.56$ & $+0.39$ & $+0.33$ \\
    \midrule
    Rare tokens, first mention
      & 35,323 & $5.79$ & $+0.06$ & $+0.04$ & $0.00$ & $-0.06$ \\
    \midrule
    \multicolumn{7}{@{}l}{\emph{Rare tokens, second mention with a new prefix}} \\
    \quad 1--511 tokens back
      & 16,158 & $4.03$ & $+0.14$ & $+0.07$ & $+0.04$ & $-0.11$ \\
    \quad 512+ tokens back
      & 3,541 & $5.36$ & $+0.15$ & $+0.20$ & $+0.13$ & $+0.02$ \\
    \midrule
    \multicolumn{7}{@{}l}{\emph{Rare tokens, second mention with the same prefix}} \\
    \quad 1--511 tokens back
      & 12,472 & $1.79$ & $+0.22$ & $+0.24$ & $+0.18$ & $+0.13$ \\
    \quad 512+ tokens back
      & 3,158 & $2.41$ & $+0.66$ & $+0.85$ & $+0.63$ & $+0.52$ \\
    \bottomrule
  \end{tabular}
  \caption{
  Prediction loss on held-out FineWeb-Edu, in nats.
Softmax losses are absolute; linear-model losses are differences from softmax. For second mentions, prefix denotes the preceding token and distance is from the first mention.}
  \label{tab:zoo-copyk-fineweb}
\end{table}

Starting from the pretraining distribution, we evaluate how well the models utilize context on a held-out sample of 16M tokens. Following Zoology \citep{arora2024zoology}, we define an AR-hit as the second token in a bigram which has already appeared earlier in the same context and whose estimated corpus frequency is at most 1,250 occurrences per 10B tokens. The earlier occurrence allows the model to retrieve the completion
from context, while restricting the analysis to rare bigrams makes it less likely that the association was memorized during training. With this definition, we find that 1.7\% of tokens are AR-hits.\footnote{Zoology reports 6.4\%. Differences in corpus composition and tokenization contribute to this discrepancy; \autoref{app:zoology_data} provides a detailed comparison.}

On AR-hits, the fixed-state models incur $0.12$--$0.22$ nats higher loss
than softmax attention (\autoref{tab:zoo-copyk-fineweb}). On the remaining
tokens, the differences range from $-0.02$ to $+0.04$ nats. All four
fixed-state models thus show a larger deficit on repeated rare bigrams.

To examine the role of the cue, we also track the first and second
occurrences of rare tokens within each context, applying the same
corpus-frequency threshold to individual tokens. We group second mentions
by their distance from the first and by whether the immediately preceding
token matches across mentions. At the first mention, all models have
similarly high loss. At the second mention, every model has lower loss
with a matching prefix in both distance bins.

The largest gap occurs at distant second mentions with a matching prefix. When the first mention is at least 512 tokens back, fixed-state losses are $0.02$--$0.20$ nats higher than softmax with a new prefix and
$0.52$--$0.85$ nats higher with the same prefix. At shorter distances,
the same-prefix gap is $0.13$--$0.24$ nats. Softmax attention therefore
shows its clearest advantage when the preceding token supplies an exact
cue to a distant mention.

\FloatBarrier
\subsection{Recall Benchmark Suite}

To evaluate how the gap affects model behavior, we use a standard suite of six benchmarks \citep{arora2024justreadtwiceclosing}. SWDE \citep{swde,arora2025languagemodelsenablesimple} and FDA \citep{Wu2021MedicalAI} measure attribute extraction from documents. SQuAD \citep{rajpurkar2016squad100000questionsmachine}, TriviaQA \citep{tqa}, Natural Questions \citep{nq}, and DROP \citep{dua2019dropreadingcomprehensionbenchmark} use cloze-formatted prompts with supporting documents in context. In the notebook example above, the original question asks where Maya left it, while the prefix prompt supplies ``her notebook in the'' and asks the model to continue with ``kitchen''.

\begin{figure}[!htbp]
  \centering
  \begingroup

\definecolor{figlinear}{HTML}{1F77B4}
\definecolor{figsoftmax}{HTML}{2B2B2B}
\definecolor{figkda}{HTML}{92702D}

\newcommand{\figonefont}{%
  \fontencoding{T1}\fontfamily{DejaVuSans-TLF}\selectfont
}

\newcommand{\figonebar}[3]{%
  \path[fill=#3]
    (axis cs:#1,0) rectangle
    (axis cs:{#1+0.62},#2);
  \node[
    anchor=south,
    text=white,
    inner sep=0pt,
    font=\figonefont\fontsize{6.5}{8}\selectfont
  ] at (axis cs:{#1+0.31},1.5) {%
    \pgfmathprintnumber[
      fixed,
      fixed zerofill,
      precision=1
    ]{#2}%
  };
}

\newcommand{\figonepplbar}[4]{%
  \path[fill=#3]
    (axis cs:#1,#4) rectangle
    (axis cs:{#1+0.62},#2);
  \node[
    anchor=south,
    text=white,
    inner sep=0pt,
    font=\figonefont\fontsize{6.5}{8}\selectfont
  ] at (axis cs:{#1+0.31},{#4+0.12}) {%
    \pgfmathprintnumber[
      fixed,
      fixed zerofill,
      precision=1
    ]{#2}%
  };
}

\begin{tikzpicture}[x=1in,y=1in]
\path[use as bounding box] (0,0) rectangle (5.5,2.55);

\begin{axis}[
  at={(0.49in,0.64in)},
  anchor=south west,
  width=1.70in,
  height=1.60in,
  scale only axis,
  xmin=-0.25,
  xmax=5.45,
  ymin=0,
  ymax=100,
  axis lines=left,
  axis line style={draw=black!75,line width=.55pt},
  ymajorgrids,
  grid style={draw=black!13,line width=.4pt},
  tick align=outside,
  scaled ticks=false,
  xtick={0.31,1.11,1.91,3.31,4.11,4.91},
  xticklabels={Lin.,Soft.,KDA,Lin.,Soft.,KDA},
  ytick={0,20,40,60,80,100},
  tick label style={
    font=\figonefont\fontsize{6.5}{8}\selectfont
  },
  label style={
    font=\figonefont\fontsize{7.5}{9}\selectfont
  },
  title style={
    yshift=3pt,
    font=\figonefont\fontsize{8}{10}\selectfont
  },
  title={Document extraction},
  ylabel={Accuracy (\%)},
  clip=false
]

\figonebar{0.00}{15.6875}{figlinear}
\figonebar{0.80}{52.1}{figsoftmax}
\figonebar{1.60}{21.0}{figkda}

\figonebar{3.00}{19.65}{figlinear}
\figonebar{3.80}{67.65}{figsoftmax}
\figonebar{4.60}{25.5}{figkda}

\node[
  anchor=north,
  font=\figonefont\fontsize{6.5}{8}\selectfont
] at (axis cs:1.11,-15) {Base};

\node[
  anchor=north,
  font=\figonefont\fontsize{6.5}{8}\selectfont
] at (axis cs:4.11,-15) {Prefix};

\end{axis}

\begin{axis}[
  at={(2.50in,0.64in)},
  anchor=south west,
  width=1.70in,
  height=1.60in,
  scale only axis,
  xmin=-0.25,
  xmax=5.45,
  ymin=0,
  ymax=100,
  axis lines=left,
  axis line style={draw=black!75,line width=.55pt},
  ymajorgrids,
  grid style={draw=black!13,line width=.4pt},
  tick align=outside,
  scaled ticks=false,
  xtick={0.31,1.11,1.91,3.31,4.11,4.91},
  xticklabels={Lin.,Soft.,KDA,Lin.,Soft.,KDA},
  ytick={0,20,40,60,80,100},
  tick label style={
    font=\figonefont\fontsize{6.5}{8}\selectfont
  },
  title style={
    yshift=3pt,
    font=\figonefont\fontsize{8}{10}\selectfont
  },
  title={In-context QA},
  clip=false
]

\figonebar{0.00}{27.91875}{figlinear}
\figonebar{0.80}{29.275}{figsoftmax}
\figonebar{1.60}{30.2}{figkda}

\figonebar{3.00}{49.30625}{figlinear}
\figonebar{3.80}{68.65}{figsoftmax}
\figonebar{4.60}{52.625}{figkda}

\node[
  anchor=north,
  font=\figonefont\fontsize{6.5}{8}\selectfont
] at (axis cs:1.11,-15) {Base};

\node[
  anchor=north,
  font=\figonefont\fontsize{6.5}{8}\selectfont
] at (axis cs:4.11,-15) {Prefix};

\end{axis}

\begin{axis}[
  at={(4.57in,0.64in)},
  anchor=south west,
  width=0.85in,
  height=1.60in,
  scale only axis,
  xmin=-0.25,
  xmax=2.45,
  ymin=25,
  ymax=37.5,
  axis lines=left,
  axis line style={draw=black!75,line width=.55pt},
  ymajorgrids,
  grid style={draw=black!13,line width=.4pt},
  tick align=outside,
  scaled ticks=false,
  xtick={0.31,1.11,1.91},
  xticklabels={Lin.,Soft.,KDA},
  ytick={26,28,30,32,34,36},
  tick label style={
    font=\figonefont\fontsize{6.5}{8}\selectfont
  },
  title style={
    yshift=3pt,
    font=\figonefont\fontsize{8}{10}\selectfont
  },
  title={LM perplexity},
  clip=false
]

\figonepplbar{0.00}{29.975}{figlinear}{25}
\figonepplbar{0.80}{30.65}{figsoftmax}{25}
\figonepplbar{1.60}{27.8}{figkda}{25}

\end{axis}

\node[
  anchor=south,
  align=center,
  inner sep=0pt,
  font=\figonefont\fontsize{7}{8.4}\selectfont
] at (2.75,0.02) {%
  \tikz[baseline=-.5ex]{
    \fill[figlinear] (0,0) rectangle (8pt,5pt);
  }\,Linear mean
  \qquad
  \tikz[baseline=-.5ex]{
    \fill[figsoftmax] (0,0) rectangle (8pt,5pt);
  }\,Softmax
  \qquad
  \tikz[baseline=-.5ex]{
    \fill[figkda] (0,0) rectangle (8pt,5pt);
  }\,KDA
};

\end{tikzpicture}

\par
\endgroup
  \caption{Recall accuracy and language modeling perplexity under
    standard training. The first two panels report mean accuracy
    on document extraction (SWDE and FDA) and in-context QA
    (SQuAD, TQA, NQ, and DROP), each evaluated with base and prefix
    prompts. The right panel reports mean perplexity on LAMBADA and
    WikiText. Blue bars show the mean across linear-attention
    architectures, black bars show softmax, and ochre bars show KDA
    separately. Numbers inside the bars give the scores. Higher
    accuracy and lower perplexity are better.}
  \label{fig:recall-suite}
\end{figure}
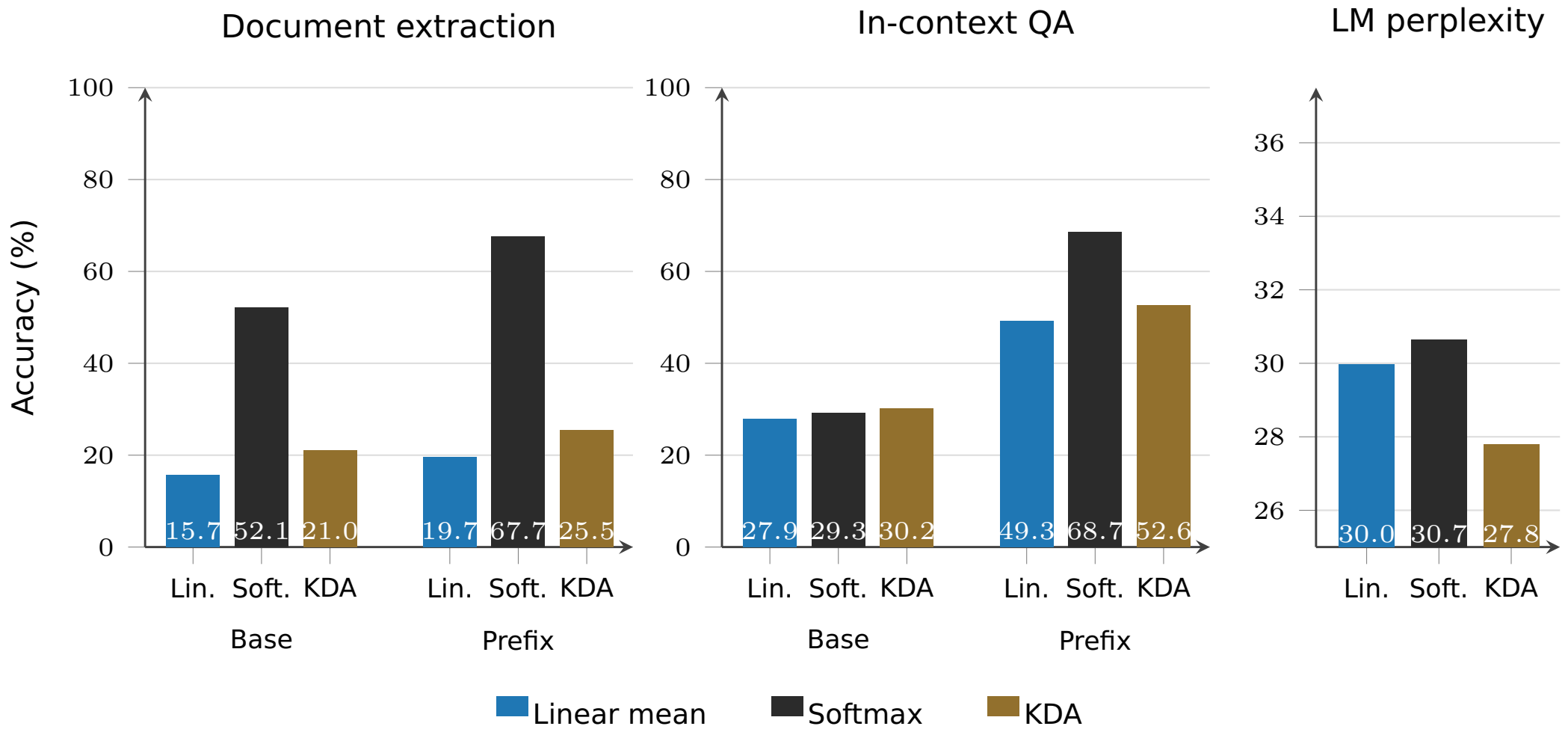

For these tasks, softmax attention is clearly better for document extraction, while the gap on QA is small. Different linear attention variants perform relatively similarly, with KDA the strongest overall performer. Unlike the token-level analysis above, these benchmarks measure aggregate answer accuracy, which depends on interpreting the question and generating the answer as well as retrieving information from context. A small gap on the original QA prompts therefore need not imply that the models benefit equally from earlier context.

One relevant difference between the benchmarks is how the generation prompt is formatted: for SWDE and FDA, the cues closely resemble how the information is initially presented, while the QA tasks use a cloze \citep{Taylor1953Cloze,Brown2020LanguageModelsFewShotLearners} format that phrases the prompt as a reply to the question with the final answer left out \citep{arora2024justreadtwiceclosing}. These prompts do not necessarily reproduce the prefix preceding the answer in context.

To test the effect of an explicit matching cue, we replace the original generation prompt with the four words immediately preceding the selected answer's first occurrence in context. Consistent with the rare-token results, these prefix prompts substantially improve accuracy, especially on the QA datasets, and expose a larger advantage for softmax attention. On the TriviaQA substitution subset, softmax generated the replacement answer on 16.4\% of examples with the original prompt and 73.8\% with the four-word prefix. For KDA, the corresponding accuracies were 14.4\% and 60.3\%. The two models thus appear closely matched when asked the original question. Giving each model the words that introduced the answer makes both more likely to follow the edited document and reveals a larger difference between them. Complete results by task appear in \autoref{app:Complete_Recall_Suite_Results}.

Context-removal and answer-substitution controls further support this distinction (\autoref{app:example-cues}). Removing context lowers accuracy for both prompt types, while prefix prompts generally make models more likely to reproduce an answer substituted into the context. The architectural gap thus depends on how retrieval is elicited: it is small on the original QA prompts but becomes more pronounced when an exact matching cue is supplied. Together with the natural-text results, this motivates studying associative storage and retrieval under controlled conditions using MQAR.

\FloatBarrier
\section{MQAR}

The observed gap in prefix-matching recall motivates a controlled
test using multi-query associative recall (MQAR)
\citep{arora2024zoology}. The task requires a model to bind
key--value pairs encountered in context and retrieve the
corresponding value when a key appears again. Keys and values
are sampled independently, preventing the associations from
being memorized during training, while multiple queries require
the model to retain several associations within one sequence.
\autoref{app:MQAR_details} describes the task construction.

\subsection{Experiments}

We compare the four sequence mixers evaluated above, denoting channel-wise GLA by D-GLA in the synthetic tasks, alongside
simple linear attention (LA) and gated linear attention with head-wise decay (H-GLA), using four settings with different numbers of key--value pairs. The vocabulary contains 8,192 tokens,
divided equally between keys and values, and sequence length is six times the number of pairs. Each model is trained for 128 epochs of 100,000 samples. The baseline model dimension is 64 with four square heads, giving the linear-attention
models a recurrent state size of $S=h d_k d_v=1024$ entries.

\begin{figure}[htbp]
  \centering
    \input{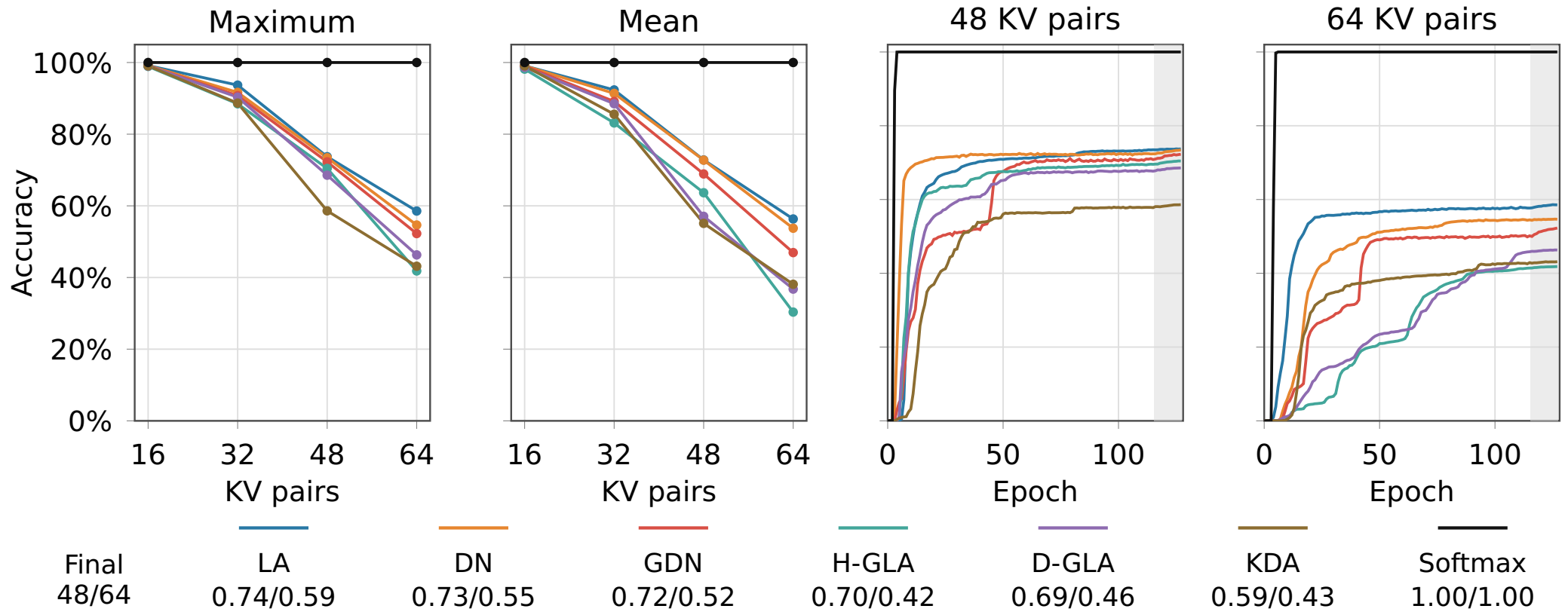}
  \caption{MQAR accuracy. The left panels show maximum and mean
    final-checkpoint accuracy across four key--value pair settings.
    Means are over three seeds at the best learning rate. The right
    panels show accuracy during training for 48 and 64 pairs. Gray
    shading marks the learning-rate cooldown.}
  \label{fig:mqar-results}
\end{figure}

Softmax attention solves the evaluated settings, while the
linear-attention variants approach similar numbers of retrieved
pairs but differ in convergence speed. LA and DN perform best,
particularly in the harder settings, while gated variants
converge more slowly and gradually approach the non-decaying
models. This agrees with earlier observations of slower
associative-recall learning in gated architectures
\citep{trockman2024mimeticinitializationhelpsstate,
okpekpe2025revisitingassociativerecallmodern}. 

\subsection{Capacity Scaling}

At state size $S=1024$, the number of successfully retrieved
pairs approaches 32. We extend the comparison to model
dimensions 32 and 128, with state sizes 256 and 4096,
respectively, scaling the number of presented pairs by the
same factor of four. LA and DN approximately follow a
relationship of $S/32$ retrieved associations at a vocabulary
size of 8,192 (\autoref{fig:mqar-capacity-scaling}).
Accuracy declines smoothly as more pairs are presented,
consistent with increasing interference rather than a sharp
capacity cutoff.
\begin{figure}[htbp]
  \centering
  \begin{minipage}[t]{0.49\linewidth}
    \vspace{0pt}
    \centering
    \begingroup
\newcommand{\synfont}{\fontencoding{T1}\fontfamily{DejaVuSans-TLF}\selectfont}
\pgfplotsset{paper synthetic/.style={scale only axis,
  tick label style={font=\synfont\fontsize{6.5}{8}\selectfont},
  label style={font=\synfont\fontsize{7}{8.4}\selectfont},
  title style={font=\synfont\fontsize{7.5}{9}\selectfont},
  axis line style={draw=black!70,line width=.45pt},
  grid style={draw=black!13,line width=.35pt},tick align=outside,
  scaled ticks=false,major tick length=2pt}}
\begin{tikzpicture}
\begin{loglogaxis}[paper synthetic,
name=LA,
  width=\dimexpr.5\linewidth-19pt\relax,height=1.14in,
  xmin=128,xmax=8192,ymin=2.5,ymax=400,
  log basis x=2,log basis y=2,xtick={256,1024,4096},xticklabels={256,1k,4k},
  ytick={4,8,16,32,64,128,256},xmajorgrids,ymajorgrids,
  yticklabels={4,8,16,32,64,128,256},ylabel={KV pairs recalled},title={Linear attention}]
\addplot[c1f77b4,dashed,mark=none,line width=.8pt] table[col sep=comma,x=x,y=y]{synth-plots/data/capacity_scaling_panel1_line10.csv};
\addplot[c1f77b4,only marks,mark=square*,mark size=1.9pt,mark options={draw=white,line width=.3pt}] table[col sep=comma,x=x,y=y]{synth-plots/data/capacity_scaling_panel1_line1.csv};
\addplot[c1f77b4,only marks,mark=*,mark size=2pt,mark options={draw=white,line width=.3pt}] table[col sep=comma,x=x,y=y]{synth-plots/data/capacity_scaling_panel1_line2.csv};
\addplot[c1f77b4,only marks,mark=synthetic star,mark size=3pt,mark options={draw=white,line width=.3pt}] table[col sep=comma,x=x,y=y]{synth-plots/data/capacity_scaling_panel1_line3.csv};
\addplot[c1f77b4,only marks,mark=square*,mark size=1.9pt,mark options={draw=white,line width=.3pt}] table[col sep=comma,x=x,y=y]{synth-plots/data/capacity_scaling_panel1_line4.csv};
\addplot[c1f77b4,only marks,mark=*,mark size=2pt,mark options={draw=white,line width=.3pt}] table[col sep=comma,x=x,y=y]{synth-plots/data/capacity_scaling_panel1_line5.csv};
\addplot[c1f77b4,only marks,mark=synthetic star,mark size=3pt,mark options={draw=white,line width=.3pt}] table[col sep=comma,x=x,y=y]{synth-plots/data/capacity_scaling_panel1_line6.csv};
\addplot[c1f77b4,only marks,mark=square*,mark size=1.9pt,mark options={draw=white,line width=.3pt}] table[col sep=comma,x=x,y=y]{synth-plots/data/capacity_scaling_panel1_line7.csv};
\addplot[c1f77b4,only marks,mark=*,mark size=2pt,mark options={draw=white,line width=.3pt}] table[col sep=comma,x=x,y=y]{synth-plots/data/capacity_scaling_panel1_line8.csv};
\addplot[c1f77b4,only marks,mark=synthetic star,mark size=3pt,mark options={draw=white,line width=.3pt}] table[col sep=comma,x=x,y=y]{synth-plots/data/capacity_scaling_panel1_line9.csv};
\node[anchor=south east,inner sep=1pt,font=\synfont\fontsize{6.5}{8}\selectfont] at (rel axis cs:.99,.02) {Fit: \textit{S}/32.14};
\end{loglogaxis}
\begin{loglogaxis}[paper synthetic,
at={(LA.south east)},anchor=south west,xshift=6pt,
  width=\dimexpr.5\linewidth-19pt\relax,height=1.14in,
  xmin=128,xmax=8192,ymin=2.5,ymax=400,
  log basis x=2,log basis y=2,xtick={256,1024,4096},xticklabels={256,1k,4k},
  ytick={4,8,16,32,64,128,256},xmajorgrids,ymajorgrids,
  yticklabels={},title={DeltaNet}]
\addplot[cff7f0e,dashed,mark=none,line width=.8pt] table[col sep=comma,x=x,y=y]{synth-plots/data/capacity_scaling_panel2_line10.csv};
\addplot[cff7f0e,only marks,mark=square*,mark size=1.9pt,mark options={draw=white,line width=.3pt}] table[col sep=comma,x=x,y=y]{synth-plots/data/capacity_scaling_panel2_line1.csv};
\addplot[cff7f0e,only marks,mark=*,mark size=2pt,mark options={draw=white,line width=.3pt}] table[col sep=comma,x=x,y=y]{synth-plots/data/capacity_scaling_panel2_line2.csv};
\addplot[cff7f0e,only marks,mark=synthetic star,mark size=3pt,mark options={draw=white,line width=.3pt}] table[col sep=comma,x=x,y=y]{synth-plots/data/capacity_scaling_panel2_line3.csv};
\addplot[cff7f0e,only marks,mark=square*,mark size=1.9pt,mark options={draw=white,line width=.3pt}] table[col sep=comma,x=x,y=y]{synth-plots/data/capacity_scaling_panel2_line4.csv};
\addplot[cff7f0e,only marks,mark=*,mark size=2pt,mark options={draw=white,line width=.3pt}] table[col sep=comma,x=x,y=y]{synth-plots/data/capacity_scaling_panel2_line5.csv};
\addplot[cff7f0e,only marks,mark=synthetic star,mark size=3pt,mark options={draw=white,line width=.3pt}] table[col sep=comma,x=x,y=y]{synth-plots/data/capacity_scaling_panel2_line6.csv};
\addplot[cff7f0e,only marks,mark=square*,mark size=1.9pt,mark options={draw=white,line width=.3pt}] table[col sep=comma,x=x,y=y]{synth-plots/data/capacity_scaling_panel2_line7.csv};
\addplot[cff7f0e,only marks,mark=*,mark size=2pt,mark options={draw=white,line width=.3pt}] table[col sep=comma,x=x,y=y]{synth-plots/data/capacity_scaling_panel2_line8.csv};
\addplot[cff7f0e,only marks,mark=synthetic star,mark size=3pt,mark options={draw=white,line width=.3pt}] table[col sep=comma,x=x,y=y]{synth-plots/data/capacity_scaling_panel2_line9.csv};
\node[anchor=south east,inner sep=1pt,font=\synfont\fontsize{6.5}{8}\selectfont] at (rel axis cs:.99,.02) {Fit: \textit{S}/32.28};
\end{loglogaxis}
\path (LA.south west) -- (current axis.south east) coordinate[midway] (capmid);
\node[anchor=north,font=\synfont\fontsize{7}{8.4}\selectfont] at ([yshift=-12pt]capmid)
  {State size (\textit{S} = heads \texttimes{} \textit{d}\textsubscript{k} \texttimes{} \textit{d}\textsubscript{v})};
\newcommand{\loadmark}[1]{\tikz[baseline=-.5ex]{\color{black!65}\pgfsetplotmarksize{2pt}\pgfuseplotmark{#1}}}
\node[anchor=north,font=\synfont\fontsize{6.5}{8}\selectfont] at ([yshift=-23pt]capmid)
  {Training load: \loadmark{square*}\,\textit{S}/32\quad \loadmark{*}\,3\textit{S}/64\quad \loadmark{synthetic star}\,\textit{S}/16};
\end{tikzpicture}
\endgroup
    \caption{Capacity scaling for LA and DN at three state
    sizes. Each point averages five seeds across four
    learning rates.}
    \label{fig:mqar-capacity-scaling}
  \end{minipage}\hfill
  \begin{minipage}[t]{0.49\linewidth}
    \vspace{0pt}
    \centering
    \input{synth-plots/plots/capacity_vs_params}
    \caption{LA recall across parameter counts and head
    layouts at three state sizes. Points average available
    seeds per configuration; dashed lines show $S/32.14$.}
    \label{fig:mqar-head-layouts}
  \end{minipage}
\end{figure}
The same scaling largely holds across a broad range of parameter counts and head layouts
(\autoref{fig:mqar-head-layouts}). Using LA, we vary model width from 32 to 512, total parameters from 0.27M to 4.4M, and head count from 1 to 16, including key/value dimension ratios from 1:4 to 4:1. Projection dimensions are adjusted to keep state size fixed at 256, 1024, or 4096 entries, with each configuration trained on $S/16$ pairs. Most configurations remain close to the same relationship; the largest departures occur in narrow models at the largest state size and some
asymmetric layouts with few heads. Across this range, recall remains primarily determined by state size, with parameter count and head layout having a smaller effect. Varying width and parameter count while holding the recurrent state fixed separates representational capacity from state memory capacity: MQAR retrieval scales primarily with the latter.

We find that LA’s recall capacity grows approximately linearly with recurrent state size \(S\), reaching \(S/32\) associations at a vocabulary size of \(8{,}192\) and \(S/36\) at \(32{,}768\). This scaling is consistent with the dependence on state and
vocabulary size reported for Mamba by
\citet{koren2026recallscalinglawsmamba}. Our results thus suggest that the same scaling holds for linear attention models.

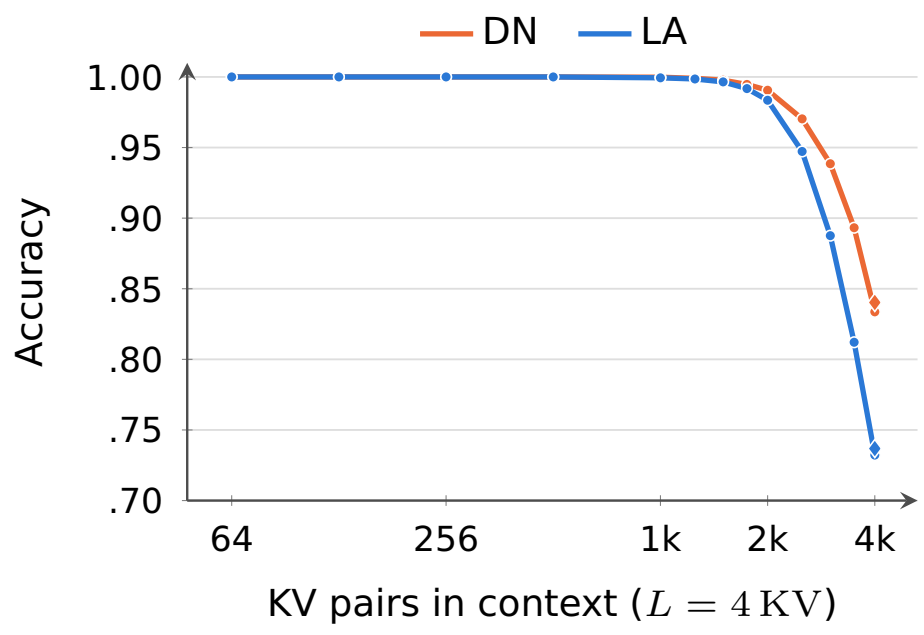
\begin{wrapfigure}[13]{r}{0.40\textwidth}
  \centering
\begingroup
\newcommand{\synfont}{\fontencoding{T1}\fontfamily{DejaVuSans-TLF}\selectfont}
\pgfplotsset{paper synthetic/.style={synthetic axis,scale only axis,
  tick label style={font=\synfont\fontsize{6.5}{8}\selectfont},
  label style={font=\synfont\fontsize{7}{8.4}\selectfont},
  title style={font=\synfont\fontsize{7.5}{9}\selectfont},
  axis line style={draw=black!70,line width=.45pt},
  grid style={draw=black!13,line width=.35pt},tick align=outside,
  scaled ticks=false,major tick length=2pt}}
\begin{tikzpicture}
\begin{axis}[paper synthetic,name=ramp,
 width=\dimexpr\linewidth-36pt\relax,height=1.14in,axis lines=left,
 xmin=48,xmax=5400,ymin=.70,ymax=1.01,xmode=log,log basis x=2,
 xtick={64,256,1024,2048,4096},xticklabels={64,256,1k,2k,4k},
 ytick={.70,.75,.80,.85,.90,.95,1},yticklabels={.70,.75,.80,.85,.90,.95,1.00},
 xlabel={KV pairs in context ($L=4\,\mathrm{KV}$)},ylabel={Accuracy},
 ymajorgrids,xmajorgrids=false]
\addplot[ceb6834,line width=1pt,mark=*,mark size=1pt,mark options={draw=white,line width=.3pt}] table[col sep=comma,x=x,y=y]{synth-plots/data/mqar_huge_ramp_wrap_panel1_line1.csv};
\addplot[c2a78d6,line width=1pt,mark=*,mark size=1pt,mark options={draw=white,line width=.3pt}] table[col sep=comma,x=x,y=y]{synth-plots/data/mqar_huge_ramp_wrap_panel1_line3.csv};
\addplot[ceb6834,line width=1pt,only marks,mark=diamond*,mark size=1.7pt,mark options={draw=white,line width=.3pt}] table[col sep=comma,x=x,y=y]{synth-plots/data/mqar_huge_ramp_wrap_panel1_line2.csv};
\addplot[c2a78d6,line width=1pt,only marks,mark=diamond*,mark size=1.7pt,mark options={draw=white,line width=.3pt}] table[col sep=comma,x=x,y=y]{synth-plots/data/mqar_huge_ramp_wrap_panel1_line4.csv};
\end{axis}
\node[anchor=south,font=\synfont\fontsize{7}{8.4}\selectfont] at (ramp.north)
  {\tikz[baseline=-.5ex]{\draw[ceb6834,line width=1pt](0,0)--(10pt,0);}\,DN\quad
   \tikz[baseline=-.5ex]{\draw[c2a78d6,line width=1pt](0,0)--(10pt,0);}\,LA};
\end{tikzpicture}
\endgroup
  \caption{Single-layer MQAR accuracy as the number of KV pairs
    in context increases. At 4,096 pairs, DN reaches 84.0\%
    accuracy and LA reaches 73.7\%.}
  \label{fig:mqar-ramp}
\end{wrapfigure}

Using the observed \(S/36\) ratio, we predict that a single layer with the recurrent state size used in our language models can retrieve roughly \(3{,}600\) associations at the \(32{,}000\)-token Llama-2 vocabulary size.

We test this prediction by gradually increasing the number of key–value pairs during training. With 4,096 pairs in a 16k-token context, DN retrieves 3,442 and LA retrieves 3,018. At the state size used by one language model layer, dedicated recall training therefore provides a concrete reference point for how much information that layer can learn to retrieve.
\FloatBarrier
\section{Copying}

We next test whether the MQAR scaling extends to copying random tokens and natural text, using a task adapted from \citet{jelassi2024repeatmetransformersbetter}.
The model receives a BOS token, the source sequence, and a copy marker, then reproduces the source under teacher forcing. During reproduction, recurrent-state writes, erasure and decay are disabled, and loss and accuracy are measured only on the copied tokens.

Sequences are sampled uniformly with replacement from a vocabulary of 4,096 tokens, matching the value vocabulary in MQAR but without separate keys. We compare six linear-attention variants at model dimensions 32, 64, 128 and 256, with copied lengths ranging from $S/32$ to $S/16$. The number of correctly predicted tokens scales approximately as $S/30$, closely matching the $S/32$ relationship observed on MQAR. The variants achieve similar total recall at equal state size, within approximately 6\% of LA’s recall, although they differ in which positions are retained. DN, GDN and KDA favor recently stored tokens and copy the last 16 tokens 1.41× more accurately than the first 16, while LA and D-GLA show the opposite (0.88×).
Scaling to model dimension 1024 and using the same 32,000-token vocabulary size as in our language modeling experiments shows that single-layer models can correctly predict thousands of copied tokens
(\hyperref[tab:d1024-retrieval-c4096]{Table~\ref*{tab:d1024-retrieval-c4096}}).
With 4,096 random tokens to copy, LA and DN achieve
approximately 91\% and 90\% teacher-forced accuracy,
corresponding to 3,700 correctly predicted tokens.
The diagonal-gated variants D-GLA and KDA fall further
behind, despite having the same state size. Unlike the head-wise gated variants, they fail to learn near-unit retention in a substantial fraction of channels, repeatedly erasing stored information and reducing the memory available for copying (\autoref{app:copy-gate-probe}). 

\begin{figure}[!htbp]
  \centering

  \noindent
  \begin{minipage}[c]{0.37\linewidth}
    \centering
    \small
    \begin{tabular}{@{}lrr@{}}
      \toprule
      Arch. & Acc. (\%) & No. retrieved \\
      \midrule
      LA    & 91.4 & 3,743 \\
      DN    & 90.4 & 3,701 \\
      GDN   & 88.0 & 3,605 \\
      H-GLA & 83.7 & 3,428 \\
      D-GLA & 66.8 & 2,738 \\
      KDA   & 48.5 & 1,988 \\
      \bottomrule
    \end{tabular}
  \end{minipage}\hfill
  \begin{minipage}[c]{0.60\linewidth}
    \centering
    \begingroup
\newcommand{\capacityfont}{\fontencoding{T1}\fontfamily{DejaVuSans-TLF}\selectfont}
\begin{tikzpicture}
\path[use as bounding box] (0,0) rectangle (3.3in,2.12in);
\begin{axis}[
synthetic axis,
at={(0.4925in,0.35248571428571435in)},
anchor=south west,
width=2.7415in,
height=1.202285714285714in,
scale only axis,
xmin=0.84999999999999998,
xmax=2.1499999999999999,
ymin=0.88,
ymax=1.72,
xlabel={},
ylabel={},
title={},
tick label style={font=\capacityfont\fontsize{7.5}{8.625}\selectfont},
label style={font=\capacityfont\fontsize{7.5}{8.625}\selectfont},
title style={font=\capacityfont\fontsize{7.5}{8.625}\selectfont},
xtick={1,1.5,2},
xticklabels={{1},{1.5},{2}},
ytick={1,1.2,1.3999999999999999,1.6000000000000001},
yticklabels={{1.0},{1.2},{1.4},{1.6}},
axis lines=left,
axis line style={-},xtick pos=bottom,ytick pos=left,
xmajorgrids=true,
grid style={draw=black!13,line width=.4pt},
ymajorgrids=true
]
\addplot[color=cbfbfbf,line width=0.56571428571428573pt,mark=none] table[col sep=comma,x=x,y=y]{synth-plots/data/text_copy_capacity_dn_dense_panel1_line1.csv};
\addplot[color=c8c8c8c,line width=0.56571428571428573pt,dotted,mark=none] table[col sep=comma,x=x,y=y]{synth-plots/data/text_copy_capacity_dn_dense_panel1_line2.csv};
\addplot[draw=none,fill=c2878a5,fill opacity=0.14999999999999999,mark=none] table[col sep=comma,x=x,y=y]{synth-plots/data/text_copy_capacity_dn_dense_panel1_band1_1.csv} \closedcycle;
\addplot[draw=none,fill=ce6862f,fill opacity=0.14999999999999999,mark=none] table[col sep=comma,x=x,y=y]{synth-plots/data/text_copy_capacity_dn_dense_panel1_band2_1.csv} \closedcycle;
\addplot[color=c2878a5,line width=1.2257142857142858pt,mark=none] table[col sep=comma,x=x,y=y]{synth-plots/data/text_copy_capacity_dn_dense_panel1_line12.csv};
\addplot[color=ce6862f,line width=1.2257142857142858pt,mark=none] table[col sep=comma,x=x,y=y]{synth-plots/data/text_copy_capacity_dn_dense_panel1_line22.csv};
\addplot[color=c2878a5,line width=0.70714285714285707pt,only marks,mark=*,mark size=2.1214285714285714pt,mark options={draw=cffffff,fill=c2878a5,fill opacity=1,line width=0.42428571428571427pt},opacity=0.84999999999999998] table[col sep=comma,x=x,y=y]{synth-plots/data/text_copy_capacity_dn_dense_panel1_line3.csv};
\addplot[color=c2878a5,line width=0.70714285714285707pt,only marks,mark=*,mark size=2.1214285714285714pt,mark options={draw=cffffff,fill=c2878a5,fill opacity=1,line width=0.42428571428571427pt},opacity=0.84999999999999998] table[col sep=comma,x=x,y=y]{synth-plots/data/text_copy_capacity_dn_dense_panel1_line4.csv};
\addplot[color=c2878a5,line width=0.70714285714285707pt,only marks,mark=*,mark size=2.1214285714285714pt,mark options={draw=cffffff,fill=c2878a5,fill opacity=1,line width=0.42428571428571427pt},opacity=0.84999999999999998] table[col sep=comma,x=x,y=y]{synth-plots/data/text_copy_capacity_dn_dense_panel1_line5.csv};
\addplot[color=c2878a5,line width=0.70714285714285707pt,only marks,mark=square*,mark size=2.1214285714285714pt,mark options={draw=cffffff,fill=c2878a5,fill opacity=1,line width=0.42428571428571427pt},opacity=0.84999999999999998] table[col sep=comma,x=x,y=y]{synth-plots/data/text_copy_capacity_dn_dense_panel1_line6.csv};
\addplot[color=c2878a5,line width=0.70714285714285707pt,only marks,mark=square*,mark size=2.1214285714285714pt,mark options={draw=cffffff,fill=c2878a5,fill opacity=1,line width=0.42428571428571427pt},opacity=0.84999999999999998] table[col sep=comma,x=x,y=y]{synth-plots/data/text_copy_capacity_dn_dense_panel1_line7.csv};
\addplot[color=c2878a5,line width=0.70714285714285707pt,only marks,mark=square*,mark size=2.1214285714285714pt,mark options={draw=cffffff,fill=c2878a5,fill opacity=1,line width=0.42428571428571427pt},opacity=0.84999999999999998] table[col sep=comma,x=x,y=y]{synth-plots/data/text_copy_capacity_dn_dense_panel1_line8.csv};
\addplot[color=c2878a5,line width=0.70714285714285707pt,only marks,mark=triangle*,mark size=2.1214285714285714pt,mark options={draw=cffffff,fill=c2878a5,fill opacity=1,line width=0.42428571428571427pt},opacity=0.84999999999999998] table[col sep=comma,x=x,y=y]{synth-plots/data/text_copy_capacity_dn_dense_panel1_line9.csv};
\addplot[color=c2878a5,line width=0.70714285714285707pt,only marks,mark=triangle*,mark size=2.1214285714285714pt,mark options={draw=cffffff,fill=c2878a5,fill opacity=1,line width=0.42428571428571427pt},opacity=0.84999999999999998] table[col sep=comma,x=x,y=y]{synth-plots/data/text_copy_capacity_dn_dense_panel1_line10.csv};
\addplot[color=c2878a5,line width=0.70714285714285707pt,only marks,mark=triangle*,mark size=2.1214285714285714pt,mark options={draw=cffffff,fill=c2878a5,fill opacity=1,line width=0.42428571428571427pt},opacity=0.84999999999999998] table[col sep=comma,x=x,y=y]{synth-plots/data/text_copy_capacity_dn_dense_panel1_line11.csv};
\addplot[color=ce6862f,line width=0.70714285714285707pt,only marks,mark=triangle*,mark size=2.1214285714285714pt,mark options={draw=cffffff,fill=ce6862f,fill opacity=1,line width=0.42428571428571427pt},opacity=0.84999999999999998] table[col sep=comma,x=x,y=y]{synth-plots/data/text_copy_capacity_dn_dense_panel1_line13.csv};
\addplot[color=ce6862f,line width=0.70714285714285707pt,only marks,mark=triangle*,mark size=2.1214285714285714pt,mark options={draw=cffffff,fill=ce6862f,fill opacity=1,line width=0.42428571428571427pt},opacity=0.84999999999999998] table[col sep=comma,x=x,y=y]{synth-plots/data/text_copy_capacity_dn_dense_panel1_line14.csv};
\addplot[color=ce6862f,line width=0.70714285714285707pt,only marks,mark=triangle*,mark size=2.1214285714285714pt,mark options={draw=cffffff,fill=ce6862f,fill opacity=1,line width=0.42428571428571427pt},opacity=0.84999999999999998] table[col sep=comma,x=x,y=y]{synth-plots/data/text_copy_capacity_dn_dense_panel1_line15.csv};
\addplot[color=ce6862f,line width=0.70714285714285707pt,only marks,mark=square*,mark size=2.1214285714285714pt,mark options={draw=cffffff,fill=ce6862f,fill opacity=1,line width=0.42428571428571427pt},opacity=0.84999999999999998] table[col sep=comma,x=x,y=y]{synth-plots/data/text_copy_capacity_dn_dense_panel1_line16.csv};
\addplot[color=ce6862f,line width=0.70714285714285707pt,only marks,mark=square*,mark size=2.1214285714285714pt,mark options={draw=cffffff,fill=ce6862f,fill opacity=1,line width=0.42428571428571427pt},opacity=0.84999999999999998] table[col sep=comma,x=x,y=y]{synth-plots/data/text_copy_capacity_dn_dense_panel1_line17.csv};
\addplot[color=ce6862f,line width=0.70714285714285707pt,only marks,mark=square*,mark size=2.1214285714285714pt,mark options={draw=cffffff,fill=ce6862f,fill opacity=1,line width=0.42428571428571427pt},opacity=0.84999999999999998] table[col sep=comma,x=x,y=y]{synth-plots/data/text_copy_capacity_dn_dense_panel1_line18.csv};
\addplot[color=ce6862f,line width=0.70714285714285707pt,only marks,mark=*,mark size=2.1214285714285714pt,mark options={draw=cffffff,fill=ce6862f,fill opacity=1,line width=0.42428571428571427pt},opacity=0.84999999999999998] table[col sep=comma,x=x,y=y]{synth-plots/data/text_copy_capacity_dn_dense_panel1_line19.csv};
\addplot[color=ce6862f,line width=0.70714285714285707pt,only marks,mark=*,mark size=2.1214285714285714pt,mark options={draw=cffffff,fill=ce6862f,fill opacity=1,line width=0.42428571428571427pt},opacity=0.84999999999999998] table[col sep=comma,x=x,y=y]{synth-plots/data/text_copy_capacity_dn_dense_panel1_line20.csv};
\addplot[color=ce6862f,line width=0.70714285714285707pt,only marks,mark=*,mark size=2.1214285714285714pt,mark options={draw=cffffff,fill=ce6862f,fill opacity=1,line width=0.42428571428571427pt},opacity=0.84999999999999998] table[col sep=comma,x=x,y=y]{synth-plots/data/text_copy_capacity_dn_dense_panel1_line21.csv};
\draw[<->,draw=c4c4c4c] (axis description cs:0.83846153846153859,0.26671006944444431) -- (axis description cs:0.83846153846153859,0.80257936507936478);
\node[anchor=east,text=c333333,align=center,font=\capacityfont\fontsize{7.5}{8.625}\selectfont] at (axis description cs:0.82307692307692304,0.53464471726190466) {+47\%};
\draw[<->,draw=c8c8c8c] (axis description cs:0.52692307692307683,0.32927207341269848) -- (axis description cs:0.52692307692307683,0.52132936507936467);
\node[anchor=west,text=c737373,align=center,font=\capacityfont\fontsize{7.5}{8.625}\selectfont] at (axis description cs:0.53846153846153855,0.42530071924603163) {+20\%};
\end{axis}
\node[anchor=center,align=center,rotate=0.0,font=\capacityfont\fontsize{7.5}{8.625}\selectfont] at (1.8632499999999999in,0.09439047619047623in) {Requested copy length};
\node[anchor=center,align=center,rotate=90.0,font=\capacityfont\fontsize{7.5}{8.625}\selectfont] at (0.1391761904761904in,0.9536285714285715in) {Tokens copied correctly};

\newcommand{\capline}[2]{%
  \tikz[baseline=-.5ex]{\draw[#1] (0,0)--(11pt,0);}%
  \hspace{3pt}#2}
\newcommand{\capmark}[1]{%
  \tikz[baseline=-.5ex]{%
    \pgfsetplotmarksize{2.1pt}%
    \color{c737373}\pgfuseplotmark{#1}}}
\node[anchor=south,align=center,inner sep=0pt,
      font=\capacityfont\fontsize{7}{8.5}\selectfont]
  at (1.65in,1.62in) {%
    \capline{c2878a5,line width=1.22pt}{Random tokens (mean)}\quad
    \capline{ce6862f,line width=1.22pt}{FineWeb-Edu text (mean)}\\[2pt]
    \capline{cbfbfbf,line width=.57pt}{Perfect copy}\quad
    \capline{c8c8c8c,dotted,line width=.57pt}{\textsl{S}/32 reference}\\[3pt]
    State size (\textsl{S})\quad
    \capmark{*}\,256\quad
    \capmark{square*}\,1,024\quad
    \capmark{triangle*}\,4,096%
  };
\end{tikzpicture}
\endgroup
  \end{minipage}

  \par\nointerlineskip
  \noindent
  \begin{minipage}[t]{0.37\linewidth}
    \vspace{0pt}
    
      \captionof{table}{Teacher-forced copying of 4,096 random tokens
    by single-layer models with \(d=1{,}024\), \(S=131{,}072\),
    and a vocabulary of 32,000.}
  \label{tab:d1024-retrieval-c4096}
  \end{minipage}\hfill
  \begin{minipage}[t]{0.60\linewidth}
    \vspace{0pt}
    \caption{DN copying on FineWeb-Edu text and random tokens. On both axes, 1× corresponds to one token per 32 state floats. The diagonal indicates perfect copying. Lines show means across state sizes, bands show ranges, and markers show individual results.} \label{fig:copy_scaling}
  \end{minipage}
\end{figure}

We also compare copying FineWeb-Edu text with uniformly sampled sequences from the full 32,000-token Llama-2 vocabulary at model dimensions 64 and 128. Text is copied
more accurately, while the sequence mixers remain closely matched and the state-size scaling persists. This performance is not explained by strong language modeling, as without an earlier copy, the text-copying
models achieve mean next-token accuracies of 8.1\% and 11.9\%, compared to 21.4\% and 25.7\% for models trained specifically for language.

\subsection{Copying with Language Models}

The single-layer experiments measure how much a fixed state retains when trained for copying. Given the similar retrieval scaling across the tested linear mixers, we use DeltaNet as their representative in the pretrained-model comparison, evaluating its natural-text copying against softmax attention. We evaluate both before and after copy fine-tuning on source lengths sampled between 1,024 and 4,096 tokens.

We omit the BOS token and copy marker used in the single-layer experiments, so the source and its repetition form a continuous text sequence. In those experiments, the recurrent state was frozen throughout reproduction. Here, DN processes the teacher-forced inputs with its usual state updates enabled. Under teacher forcing, the model receives the correct preceding tokens of the copy even when the source passage is absent. It can therefore predict some tokens from the wording of the passage as it unfolds. Comparing the solid and dotted curves at the same source length shows how much additional accuracy the model gains from having encountered the passage once before.

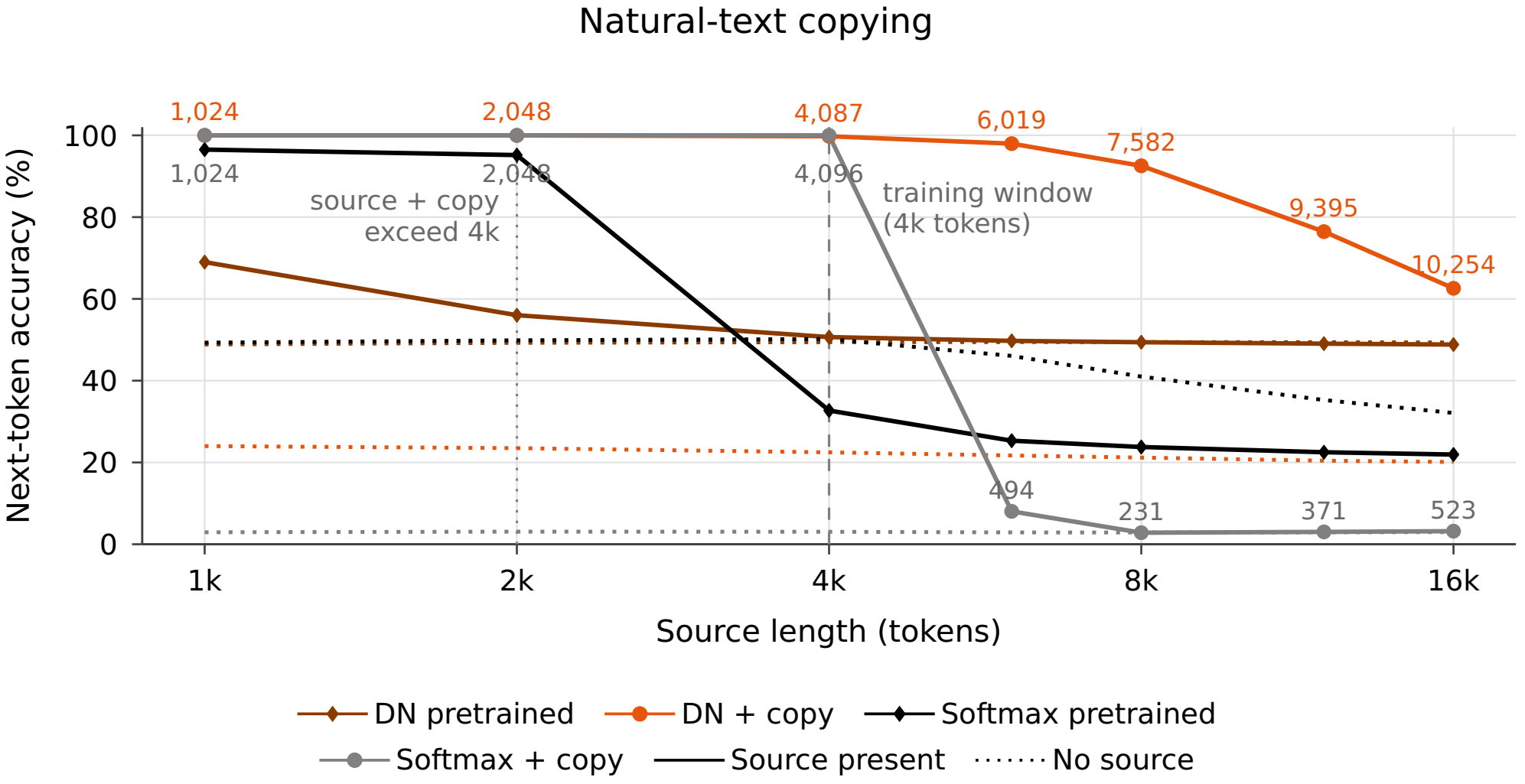
\begin{figure}[!htbp]
  \centering
  \begin{tikzpicture}[x=1in,y=1in]
\path[use as bounding box] (0,0) rectangle (5.5,2.55);

\pgfplotsset{
  copy curve/.style={
    solid,
    line width=1.1pt,
    mark size=1.65pt,
    mark options={solid,line width=.35pt}
  },
  copy baseline/.style={
    dotted,
    line width=.95pt,
    mark=none
  },
  copy guide/.style={
    color=lm808080,
    line width=.6pt,
    mark=none
  },
  copy labels/.style={
    only marks,
    mark=none,
    forget plot,
    point meta=explicit,
    nodes near coords={
      \pgfmathprintnumber[
        fixed,precision=0,1000 sep={,},assume math mode=true
      ]{\pgfplotspointmeta}
    },
    every node near coord/.append style={
      font=\copyplotfont\fontsize{6}{7}\selectfont,
      inner sep=0pt,
      outer sep=0pt
    }
  }
}

\begin{axis}[
  at={(0.66in,0.80in)},
  anchor=south west,
  width=4.675in,
  height=1.42in,
  scale only axis,
  xmin=891.44377681523156,
  xmax=18820.273848271412,
  ymin=0,
  ymax=102,
  xmode=log,
  log basis x=2,
  unbounded coords=jump,
  clip=false,
  axis lines=left,
  axis line style={-,draw=black!75,line width=.55pt},
  axis on top=false,
  tick align=outside,
  tick style={draw=black!75,line width=.5pt},
  major tick length=2.5pt,
  minor tick num=0,
  scaled ticks=false,
  xmajorgrids,
  ymajorgrids,
  grid style={draw=black!12,line width=.4pt},
  every axis plot/.append style={line join=round},
  xtick={1024,2048,4096,8192,16384},
  xticklabels={1k,2k,4k,8k,16k},
  ytick={0,20,40,60,80,100},
  yticklabels={0,20,40,60,80,100},
  tick label style={
    font=\copyplotfont\fontsize{7}{8.4}\selectfont
  },
  label style={
    font=\copyplotfont\fontsize{7.5}{9}\selectfont
  },
  xlabel={Source length (tokens)},
  ylabel={Next-token accuracy (\%)}
]

\addplot[copy guide,dotted]
  table[x=x,y=y,col sep=comma]
  {lm-plots/data/figure07_panel1_series009.csv};

\addplot[copy guide,dashed]
  table[x=x,y=y,col sep=comma]
  {lm-plots/data/figure07_panel1_series010.csv};

\addplot[copy curve,color=lm8C3B00,mark=diamond*]
  table[x=x,y=y,col sep=comma]
  {lm-plots/data/figure07_panel1_series001.csv};

\addplot[copy baseline,color=lm8C3B00]
  table[x=x,y=y,col sep=comma]
  {lm-plots/data/figure07_panel1_series002.csv};

\addplot[copy curve,color=lmE6550D,mark=*]
  table[x=x,y=y,col sep=comma]
  {lm-plots/data/figure07_panel1_series003.csv};

\addplot[copy baseline,color=lmE6550D]
  table[x=x,y=y,col sep=comma]
  {lm-plots/data/figure07_panel1_series004.csv};

\addplot[copy curve,color=lm000000,mark=diamond*]
  table[x=x,y=y,col sep=comma]
  {lm-plots/data/figure07_panel1_series005.csv};

\addplot[copy baseline,color=lm000000]
  table[x=x,y=y,col sep=comma]
  {lm-plots/data/figure07_panel1_series006.csv};

\addplot[copy curve,color=lm808080,mark=*]
  table[x=x,y=y,col sep=comma]
  {lm-plots/data/figure07_panel1_series007.csv};

\addplot[copy baseline,color=lm808080]
  table[x=x,y=y,col sep=comma]
  {lm-plots/data/figure07_panel1_series008.csv};

\def\LMcopyLeftAnnotation{source + copy\\exceed 4k}
\node[
  anchor=east, xshift=-3pt, align=right, inner sep=1pt,
  text=lm696969,
  font=\copyplotfont\fontsize{6.5}{7.6}\selectfont
] at (axis cs:2048,80) {\LMcopyLeftAnnotation};

\def\LMcopyRightAnnotation{training window\\(4k tokens)}
\node[
  anchor=west, xshift=12pt, align=left, inner sep=1pt,
  text=lm696969,
  font=\copyplotfont\fontsize{6.5}{7.6}\selectfont
] at (axis cs:4096,82) {\LMcopyRightAnnotation};

\addplot[
  copy labels,
  every node near coord/.append style={
    anchor=south,yshift=3pt,text=lmE6550D
  }
] table[
  x=x,y=y,col sep=comma,
  meta expr={round(\thisrow{x}*\thisrow{y}/100)}
] {lm-plots/data/figure07_panel1_series003.csv};

\addplot[
  copy labels,
  restrict y to domain=50:102,
  every node near coord/.append style={
    anchor=north,yshift=-7pt,text=lm696969
  }
] table[
  x=x,y=y,col sep=comma,
  meta expr={round(\thisrow{x}*\thisrow{y}/100)}
] {lm-plots/data/figure07_panel1_series007.csv};

\addplot[
  copy labels,
  restrict y to domain=0:49.99999,
  every node near coord/.append style={
    anchor=south,yshift=3pt,text=lm696969
  }
] table[
  x=x,y=y,col sep=comma,
  meta expr={round(\thisrow{x}*\thisrow{y}/100)}
] {lm-plots/data/figure07_panel1_series007.csv};

\end{axis}

\node[
  anchor=south,
  align=center,
  inner sep=0pt,
  font=\copyplotfont\fontsize{7}{8.4}\selectfont
] at (2.75,0.01) {%
  \LMkey{lm8C3B00}{solid}{diamond*}{DN pretrained}\quad
  \LMkey{lmE6550D}{solid}{*}{DN + copy}\quad
  \LMkey{lm000000}{solid}{diamond*}{Softmax pretrained}\\[3pt]
  \LMkey{lm808080}{solid}{*}{Softmax + copy}\quad
  \LMkey{lm000000}{solid}{none}{Source present}\quad
  \LMkey{lm000000}{dotted}{none}{No source}%
};

\node[
  anchor=north,
  yshift=6pt,
  inner sep=0pt,
  font=\copyplotfont\fontsize{8.5}{10}\selectfont
] at (2.75,2.55) {Natural-text copying};

\end{tikzpicture}
  \caption{Teacher-forced next-token accuracy with the source present (solid) or absent (dotted), before and after copy fine-tuning. Vertical lines mark the source lengths at which source and copy together (2,048 tokens) or the source alone (4,096 tokens) fill the pretraining window. Numbers give the count of correctly predicted copy tokens for DN + copy (orange, above markers) and softmax + copy (gray).}
  \label{fig:copying-at-scale}
\end{figure}
At a source length of 2,048 tokens, pretrained softmax attention copies with 95.4\% accuracy, compared with 56\% for DN (\autoref{fig:copying-at-scale}). At 4,096 tokens, DN falls to near its text-only baseline of 50\%. This suggests that DN gains little from the source, even though it contains the exact token sequence to be predicted. Softmax accuracy falls to 33\% as the copy extends beyond the pretraining window.

Copy fine-tuning raises DN accuracy to 99.8\% at 4,096 tokens and yields perfect copying for softmax attention across all tested lengths up to 4,096. Beyond this range, the RoPE-based softmax model falls to 3\% at 8,192 tokens, while DN declines gradually to 62\% at 16,384. Further training on source lengths of 4,096--16,384 tokens raises DN accuracy to 98\% at 16,384 tokens, with 77\% accuracy at 32,768.

Direct copying training substantially reduces next-token accuracy without the source, from 49.4\% to 22.4\%. The copying gains at fixed state size show that targeted training enables the model to use existing state memory to retain and retrieve its context.

\section{Incentivizing Recall}

MQAR and copying demonstrate that linear-attention models can retain and retrieve thousands of associations or tokens when trained directly for recall. Standard language model training, however, does not appear to reliably elicit the same behavior for linear models. This difference is consistent with architectural biases toward recent or less selective mixing, which have been theorized to harm recall performance \citep{zhang2024hedgehogporcupineexpressive,wang2025understandingmitigatingbottlenecksstate}.

While targeted recall training enables the models to learn this behavior, AR-hits account for only \(1.7\%\) of tokens in the held-out language data. The language modeling objective therefore provides relatively little direct incentive to improve this operation compared with other predictive demands. To elicit stronger recall and narrow the gap to softmax attention, we shift this balance toward retrieval by introducing repeated sentences into the training context and upweighting predictions that benefit from distant context. These interventions trade some emphasis on general language modeling for stronger recall supervision.

\begin{figure}[htbp]
  \centering
  \input{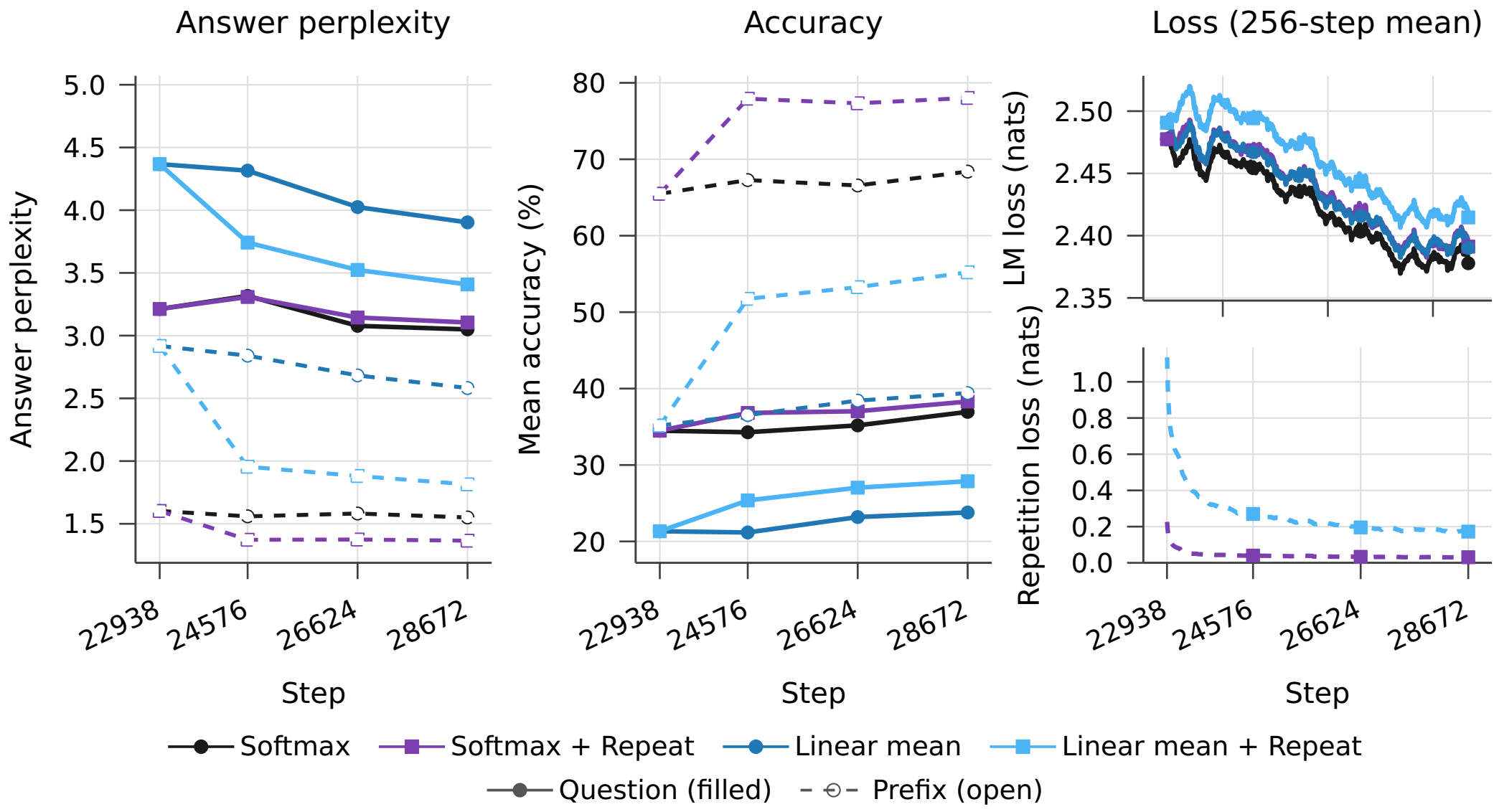}
  \caption{Answer perplexity and accuracy of the recall suite of benchmarks, evaluated at four intermediate checkpoints during the LR-cooldown phase. Loss is split into loss on regular LM tokens and loss on the repeated sequence. Linear attention aggregates the mean of each objective across the four variants.
  }
  \label{fig:AR_cooldown_phase}
\end{figure}

\subsection{Repetition In Context}

Prior work links repetition in training data to in-context learning
and induction-head emergence
\citep{chan2022datadistributionalpropertiesdrive,zucchet2025emergencesparseattentionimpact,aoyama2026predictingemergenceinductionheads}. Related approaches target recurrent-model recall through copying objectives in Birdie \citep{blouir2025birdieadvancingstatespace} and prompt repetition or non-causal prefix processing in JRT \citep{arora2024justreadtwiceclosing}. Our copying experiments show that targeted training improves retrieval but degrades language modeling when it replaces the original objective.

To better balance the trade-off, we therefore vary the proportion of repeated text within otherwise standard language modeling data, retaining causal next-token training and unchanged inference. Sentence repetition copies sentence-aligned spans averaging 64 tokens to later sentence boundaries within the same context. At a sentence repetition rate of 10\%, this increases the fraction of tokens that qualify as AR-hits to 2.86\%. Sequence repetition instead appends a full copy after the original sequence. We compare their rates in \autoref{app:recall-training-ablations}. 

\subsection{Loss Reweighting}
Alternatively, we strengthen recall incentives by reweighting the loss. We evaluate direct AR-hit upweighting and variants of LongCE (LCE) \citep{fang2025wrongperplexitylongcontextlanguage}, which emphasizes tokens whose prediction benefits from longer context. We adopt Thresholded LongCE (TLCE), a minor modification that improves the recall--language modeling tradeoff in our ablations. Short-context scores use windows of at most 512 tokens, with state and attention reset at window boundaries, and require only one additional forward pass without gradients.

Let $\Delta_i=\operatorname{sg}(\log p_{\mathrm{full}}(y_i)-\log p_{\mathrm{short}}(y_i))$, where $\operatorname{sg}$ denotes stop-gradient. TLCE uses
\begin{equation}
\mathcal{L}_{\mathrm{TLCE}}
=-\frac{\sum_i w_i\log p_{\mathrm{full}}(y_i)}{\sum_i w_i},
\qquad
w_i=
\begin{cases}
\min\{e^{\Delta_i},5\}, & \Delta_i>1 \;\text{and}\; i-s_i\ge128,\\
1, & \text{otherwise},
\end{cases}
\end{equation}
where $s_i$ is the preceding window boundary and sums include only valid target tokens. We retain LCE's capped likelihood-ratio weighting only for tokens with a context benefit exceeding one nat and at least 128 tokens of short context. All other tokens receive unit weight, and normalization by the total weight keeps the mean effective weight at one. 

Sentence repetition creates additional positions where an earlier passage supplies the continuation. TLCE can also emphasize predictions in unrepeated text, provided the model already assigns the observed token sufficiently higher probability with the longer context. The weighting rule therefore depends on what the model can do at that point in training. A distant cue that would help predict a token receives no extra weight from TLCE if the model has yet to learn to use it.

\section{Results}

Our ablations favor sentence repetition over full-sequence repetition and TLCE over LCE. On KDA, sentence repetition at 10\% achieves higher extraction and prefix accuracy than full-sequence repetition at 12.5\% (31.1\% versus 28.6\% and 58.7\% versus 56.8\%), at comparable held-out loss cost. Higher sentence-repetition rates yield diminishing recall gains while further increasing language modeling loss.

TLCE similarly retains recall performance close to LCE while reducing the held-out loss increase from 0.0867 to 0.0333 nats. We therefore select sentence repetition at 10\% and TLCE for the main experiments; full comparisons are provided in \autoref{app:recall-training-ablations}.

We apply these interventions at the start of cooldown, covering the final 20\% of training. Recall improves rapidly after their introduction, accompanied by an initial increase in language modeling loss. Recall gains subsequently slow while language modeling loss recovers toward the baseline (\autoref{fig:AR_cooldown_phase}). To assess the effect of longer exposure, we additionally train one KDA model with both sentence repetition and TLCE enabled throughout training, rather than only during cooldown.

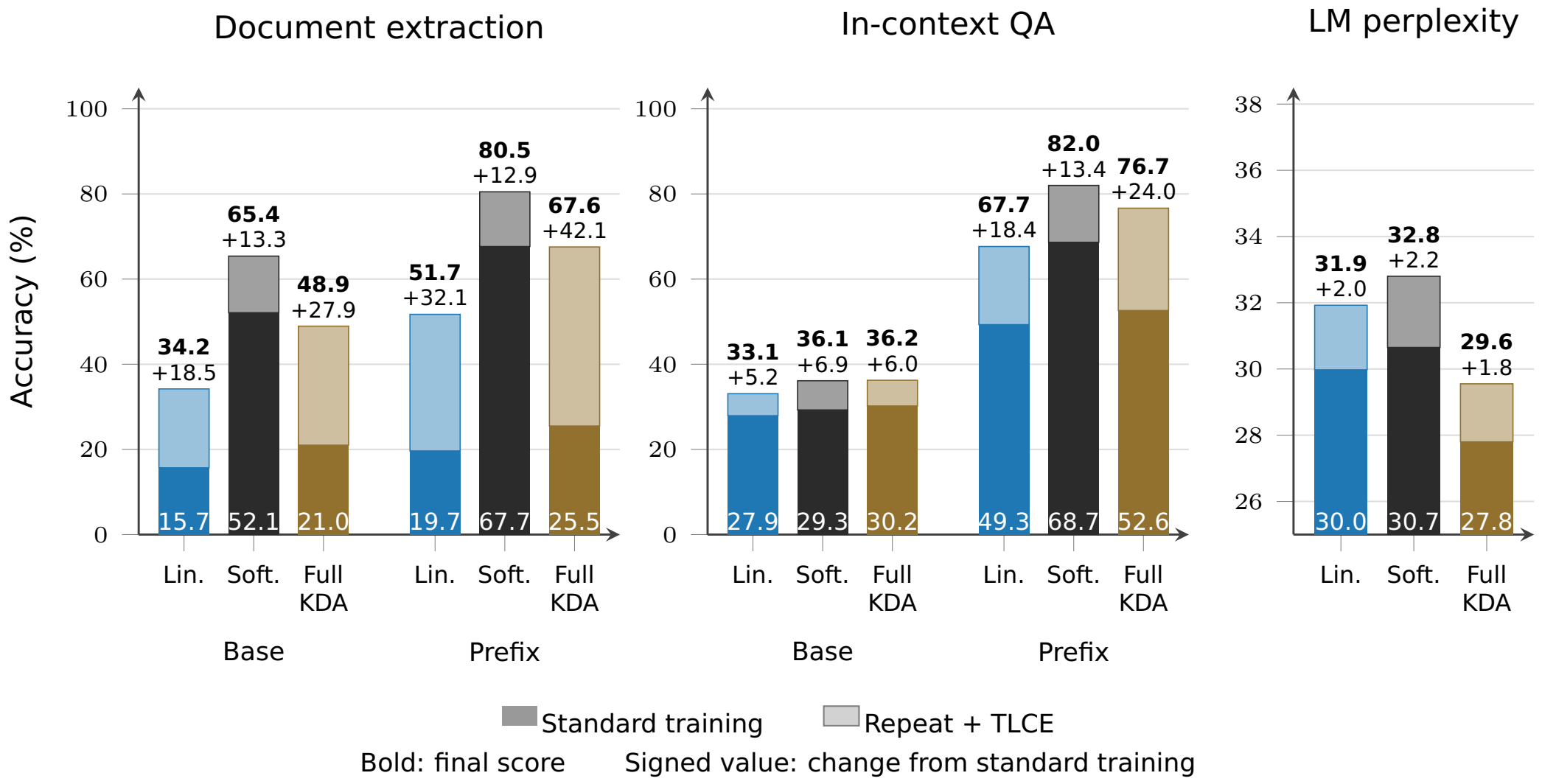
\begin{figure}[!t]
  \centering
\begingroup

\definecolor{figninelinear}{HTML}{1F77B4}
\definecolor{figninesoftmax}{HTML}{2B2B2B}
\definecolor{figninekda}{HTML}{92702D}

\newcommand{\figninefont}{%
  \fontencoding{T1}\fontfamily{DejaVuSans-TLF}\selectfont
}

\newcommand{\figninebar}[7]{%
  \path[fill=#2]
    (axis cs:{#1-0.36},#3)
    rectangle
    (axis cs:{#1+0.36},#4);

  \path[
    fill=#2!45!white,
    draw=#2,
    line width=.3pt
  ]
    (axis cs:{#1-0.36},#4)
    rectangle
    (axis cs:{#1+0.36},#5);

  \node[
    anchor=south,
    inner sep=0pt,
    yshift=1pt,
    text=white,
    font=\figninefont\fontsize{6}{7}\selectfont
  ] at (axis cs:#1,#3) {#6};

  \node[
    anchor=south,
    align=center,
    inner sep=0pt,
    yshift=2pt,
    font=\figninefont\fontsize{5.5}{6.5}\selectfont
  ] at (axis cs:#1,#5) {#7};
}

\begin{tikzpicture}[x=1in,y=1in]

\path[use as bounding box] (0,0) rectangle (5.5,2.8);

\begin{axis}[
  at={(0.49in,0.88in)},
  anchor=south west,
  width=1.70in,
  height=1.58in,
  scale only axis,
  xmin=-0.65,
  xmax=6.25,
  ymin=0,
  ymax=105,
  axis lines=left,
  axis line style={draw=black!75,line width=.55pt},
  ymajorgrids,
  grid style={draw=black!13,line width=.4pt},
  tick align=outside,
  scaled ticks=false,
  xtick={0,1,2,3.6,4.6,5.6},
  xticklabels={
    Lin.,Soft.,{Full\\KDA},
    Lin.,Soft.,{Full\\KDA}
  },
  xticklabel style={align=center},
  ytick={0,20,40,60,80,100},
  tick label style={
    font=\figninefont\fontsize{6}{7.2}\selectfont
  },
  label style={
    font=\figninefont\fontsize{7.5}{9}\selectfont
  },
  title style={
    yshift=3pt,
    font=\figninefont\fontsize{8}{10}\selectfont
  },
  title={Document extraction},
  ylabel={Accuracy (\%)},
  clip=false
]

\figninebar{0}{figninelinear}{0}{15.6875}{34.2125}
  {15.7}{\textbf{34.2}\\+18.5}

\figninebar{1}{figninesoftmax}{0}{52.1}{65.4}
  {52.1}{\textbf{65.4}\\+13.3}

\figninebar{2}{figninekda}{0}{21.0}{48.9483}
  {21.0}{\textbf{48.9}\\+27.9}

\figninebar{3.6}{figninelinear}{0}{19.65}{51.725}
  {19.7}{\textbf{51.7}\\+32.1}

\figninebar{4.6}{figninesoftmax}{0}{67.65}{80.5}
  {67.7}{\textbf{80.5}\\+12.9}

\figninebar{5.6}{figninekda}{0}{25.5}{67.5573}
  {25.5}{\textbf{67.6}\\+42.1}

\node[
  anchor=north,
  inner sep=0pt,
  font=\figninefont\fontsize{6.5}{8}\selectfont
] at (axis cs:1,-25) {Base};

\node[
  anchor=north,
  inner sep=0pt,
  font=\figninefont\fontsize{6.5}{8}\selectfont
] at (axis cs:4.6,-25) {Prefix};

\end{axis}

\begin{axis}[
  at={(2.50in,0.88in)},
  anchor=south west,
  width=1.70in,
  height=1.58in,
  scale only axis,
  xmin=-0.65,
  xmax=6.25,
  ymin=0,
  ymax=105,
  axis lines=left,
  axis line style={draw=black!75,line width=.55pt},
  ymajorgrids,
  grid style={draw=black!13,line width=.4pt},
  tick align=outside,
  scaled ticks=false,
  xtick={0,1,2,3.6,4.6,5.6},
  xticklabels={
    Lin.,Soft.,{Full\\KDA},
    Lin.,Soft.,{Full\\KDA}
  },
  xticklabel style={align=center},
  ytick={0,20,40,60,80,100},
  tick label style={
    font=\figninefont\fontsize{6}{7.2}\selectfont
  },
  title style={
    yshift=3pt,
    font=\figninefont\fontsize{8}{10}\selectfont
  },
  title={In-context QA},
  clip=false
]

\figninebar{0}{figninelinear}{0}{27.91875}{33.0875}
  {27.9}{\textbf{33.1}\\+5.2}

\figninebar{1}{figninesoftmax}{0}{29.275}{36.125}
  {29.3}{\textbf{36.1}\\+6.9}

\figninebar{2}{figninekda}{0}{30.2}{36.2437}
  {30.2}{\textbf{36.2}\\+6.0}

\figninebar{3.6}{figninelinear}{0}{49.30625}{67.65625}
  {49.3}{\textbf{67.7}\\+18.4}

\figninebar{4.6}{figninesoftmax}{0}{68.65}{82.0}
  {68.7}{\textbf{82.0}\\+13.4}

\figninebar{5.6}{figninekda}{0}{52.625}{76.6604}
  {52.6}{\textbf{76.7}\\+24.0}

\node[
  anchor=north,
  inner sep=0pt,
  font=\figninefont\fontsize{6.5}{8}\selectfont
] at (axis cs:1,-25) {Base};

\node[
  anchor=north,
  inner sep=0pt,
  font=\figninefont\fontsize{6.5}{8}\selectfont
] at (axis cs:4.6,-25) {Prefix};

\end{axis}

\begin{axis}[
  at={(4.57in,0.88in)},
  anchor=south west,
  width=0.85in,
  height=1.58in,
  scale only axis,
  xmin=-0.65,
  xmax=2.65,
  ymin=25,
  ymax=38.5,
  axis lines=left,
  axis line style={draw=black!75,line width=.55pt},
  ymajorgrids,
  grid style={draw=black!13,line width=.4pt},
  tick align=outside,
  scaled ticks=false,
  xtick={0,1,2},
  xticklabels={Lin.,Soft.,{Full\\KDA}},
  xticklabel style={align=center},
  ytick={26,28,30,32,34,36,38},
  tick label style={
    font=\figninefont\fontsize{6}{7.2}\selectfont
  },
  title style={
    yshift=3pt,
    font=\figninefont\fontsize{8}{10}\selectfont
  },
  title={LM perplexity},
  clip=false
]

\figninebar{0}{figninelinear}{25}{29.975}{31.925}
  {30.0}{\textbf{31.9}\\+2.0}

\figninebar{1}{figninesoftmax}{25}{30.65}{32.8}
  {30.7}{\textbf{32.8}\\+2.2}

\figninebar{2}{figninekda}{25}{27.8}{29.5522}
  {27.8}{\textbf{29.6}\\+1.8}

\end{axis}

\node[
  anchor=south,
  align=center,
  inner sep=0pt,
  font=\figninefont\fontsize{6.5}{8}\selectfont
] at (2.75,0.02) {%
  \tikz[baseline=-.5ex]{%
    \fill[gray!80] (0,0) rectangle (9pt,5pt);%
  }\,Standard training
  \qquad
  \tikz[baseline=-.5ex]{%
    \filldraw[fill=gray!35,draw=gray]
      (0,0) rectangle (9pt,5pt);%
  }\,Repeat + TLCE\\[2pt]
  Bold: final score
  \qquad
  Signed value: change from standard training
};

\end{tikzpicture}

\par
\endgroup
  \caption{Recall-suite accuracy and LM perplexity across training interventions. Sentence repetition and TLCE provide complementary recall gains, with their combination yielding the largest improvements. Applying both throughout KDA training further improves recall without an additional LM penalty.}
  \label{fig:recall-suite-repeat}
\end{figure}

Both interventions improve prefix-prompt recall across all evaluated architectures, with complementary benefits on the original prompts (\autoref{fig:recall-suite-repeat}). Averaged across linear mixers, repetition improves document extraction by 10.9 percentage points, while TLCE improves QA by 4.6 points. Combining them increases extraction and QA accuracy by 18.5 and 5.2 points, respectively, with larger gains under prefix prompts, at a cost of 2.0 in mean LAMBADA and WikiText perplexity.

Applying both throughout KDA training raises extraction accuracy to 48.9\%, approaching standard-trained softmax at 52.1\%, and QA accuracy to 36.2\%, comparable to softmax with both interventions. Despite displacing more fresh text, full-training KDA incurs a mean perplexity increase of 1.8, compared with 1.9 for the combined cooldown run. Together with the gains from loss reweighting on unrepeated text, this suggests that the trade-off reflects the changed training objective rather than simply reduced exposure to fresh text. The combined intervention also closes the long-distance AR-hit loss gap to standard-trained softmax for all four linear mixers (\autoref{app:AR-hits-repeat}).

\FloatBarrier
\subsection{Needle In A Haystack}

\begin{figure}[htbp]
  \centering
  \input{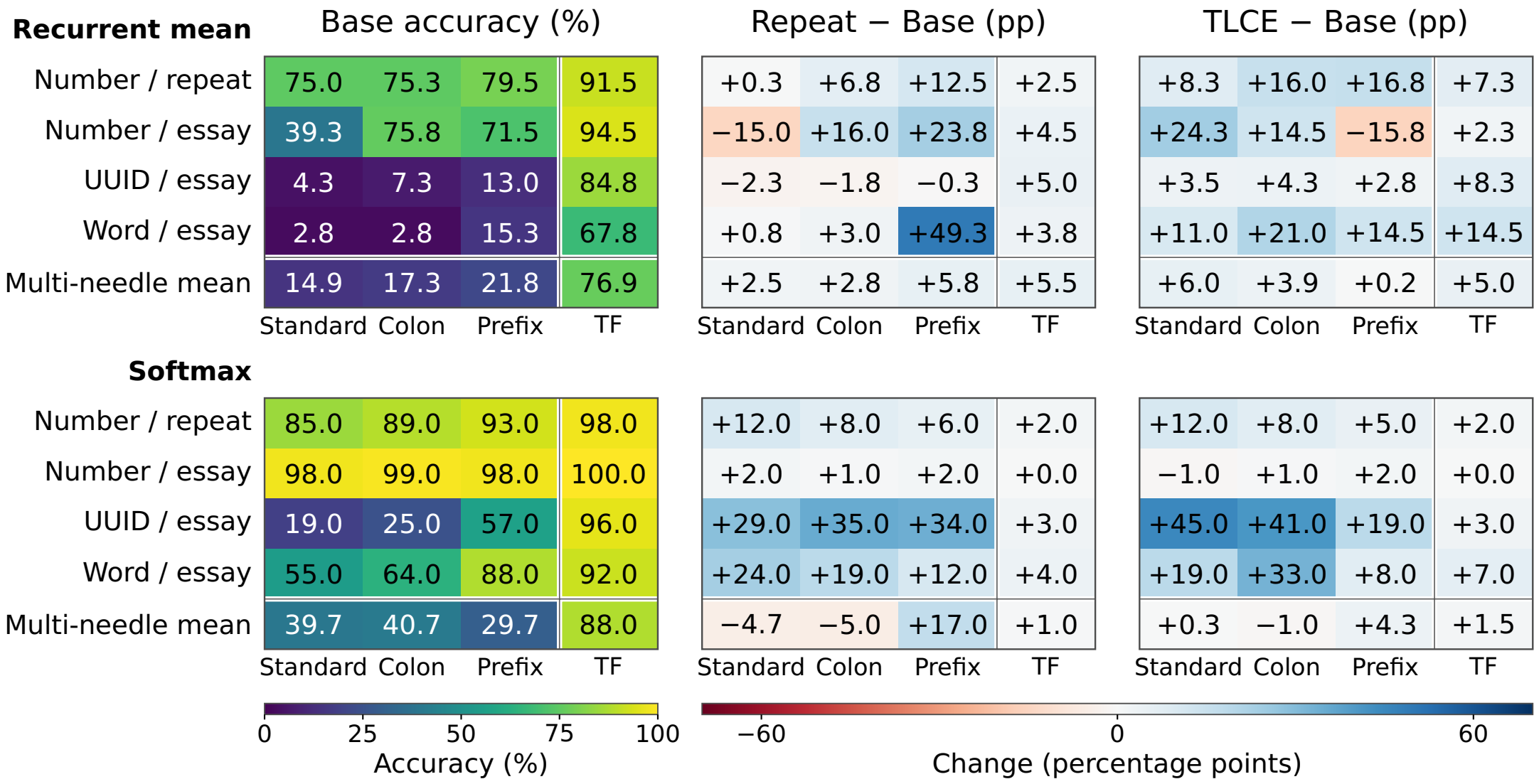}
  \caption{NIAH accuracy at 4K context length. The left column shows
    standard-training accuracy. The middle and right columns show changes
    after Repeat and TLCE training, respectively, in percentage points.
    The recurrent mean weights KDA, GDN, DN, and GLA equally. Standard,
    Colon, and Prefix report complete-answer generation accuracy. TF reports
    teacher-forced token accuracy under the Colon prompt. The multi-needle
    row averages MK-1, MQ, and MV for generation, and MK-1 and MQ for TF.}
  \label{fig:niah-main}
\end{figure}

\begin{wraptable}[13]{l}{0.4\textwidth}
    \begin{tabular}{@{}lrrrr@{}}
      \toprule
      KDA & Std. & Colon & Prefix & TF \\
      \midrule
      Base     & 37.2 & 45.9 & 53.1 & 91.9 \\
      Cooldown & 39.3 & 51.7 & 66.8 & 95.9 \\
      Full     & \textbf{54.4} & \textbf{65.8}
               & \textbf{83.8} & \textbf{98.1} \\
      \bottomrule
    \end{tabular}
   \caption{KDA accuracy (\%), averaged equally over
    2K, 4K, and 8K contexts: NIAH-1--4 for generation,
    NIAH-1--3 for teacher forcing (TF).
    Cooldown and Full apply sentence repetition and TLCE.}
    \label{tab:niah-kda-full}
\end{wraptable}

We evaluate RULER's NIAH tasks \citep{hsieh2024rulerwhatsrealcontext}:
numbers embedded in repeated text (NIAH-1), numbers or UUIDs
in natural text (NIAH-2/3), to which we introduce an additional two-word values version (NIAH-4). We compare standard,
colon-appended, and exact-prefix prompts, plus teacher-forced
token accuracy under the colon prompt. For a multi-token answer such as a UUID, teacher forcing supplies the correct preceding tokens at each step, whereas a single wrong token during generation is enough to lose the exact match.

Both Repeat and TLCE improve recall overall (\autoref{fig:niah-main}), but generation gains vary sharply across prompts, tasks, and context lengths, with localized collapses particularly on NIAH-2. Repeat yields its strongest gains on word values with exact-prefix prompts, consistent with alignment between training and evaluation cues, while transfer to other formats is less reliable. TLCE improves word retrieval across all prompt formats but also exhibits failures on other tasks. Teacher-forced results are more consistent, with both interventions improving the recurrent-model average on every single-needle task at 4K. Both also improve the multi-needle average, and applying the combined intervention throughout KDA training produces larger gains across all four measures (\autoref{tab:niah-kda-full}). Improved token-level recall nevertheless does not always translate into reliable complete-answer generation. Detailed comparisons appear in \autoref{app:NIAH_Results}.

\section{Conclusions}
Across the recall measures studied here, the clearest gap between linear and softmax attention occurs in prefix-matching retrieval, especially over long distances. In MQAR, retrieval scales primarily with recurrent state size, while increasing parameter count or changing the layout of the state at a fixed size has comparatively little effect. MQAR and copying show that models at the scale of a single language model layer can retain thousands of associations or tokens. Fixed state size therefore does not fundamentally prevent these architectures from learning effective recall, although standard language model training does not fully develop this capability.

Sentence repetition and TLCE strengthen the training signal for recall with complementary effects. Combined, they improve recall across the evaluated architectures and close the long-distance AR-hit loss gap to standard-trained softmax without increasing state size, token budget, or optimization steps. Full-training KDA further approaches baseline softmax on document extraction without a larger measured language modeling penalty than the combined cooldown intervention. These results show that linear-attention models can learn retrieval behavior comparable to that of a standard-trained softmax model while retaining their efficient inference scaling. The accompanying language modeling cost reflects a shift in training priorities that brings this capability into use. Recall-focused training also benefits softmax models, suggesting that the resulting tradeoff is relevant beyond fixed-state architectures. More broadly, the gains at unchanged state size motivate architectures that make selective, long-distance retrieval easier to learn. Improving the sharpness of retrieval and the retention of distant information offers a direction toward stronger in-context recall while preserving the favorable scaling of linear attention.
\subsection*{Acknowledgments and Disclosure of Funding}
The authors acknowledge the financial support of the Hector
Foundation. The authors also
acknowledge the Max Planck Institute for Intelligent Systems for providing the computational
resources used in this work. Antonio Orvieto is supported by the AI2050 program at Schmidt Sciences. \\
The authors thank Nadezhda Ilieva, Foivos Alimisis, and Aurelien Lucchi for the interactions and insights on recall tasks.

\subsection*{AI use statement}

In this work, we used generative AI tools to assist with implementing research code, suggesting experimental parameters, refining hypotheses, providing feedback on research methods and experiments, and interpreting and organizing results. We did not use these tools to generate synthetic datasets, clean or reformat datasets, or develop theoretical models or conceptual frameworks. Formal proofs, translation, and qualitative or thematic analysis were not part of this work.

We also used generative AI tools to create or revise figures, identify and summarize related literature, and provide feedback on the manuscript’s structure, language, and reference formatting. The authors wrote the initial complete draft and incorporated AI-assisted edits during revision. The authors reviewed AI-assisted code, approved experiments before they were run, and verified and interpreted the experimental results. The research question and overall direction originated with the authors.

The authors take responsibility for the final content of this work, including text, claims, and artifacts produced with the aid of generative AI.

\bibliographystyle{bibstyle}
\bibliography{references}

\FloatBarrier
\appendix

\section{Sequence Mixer Details}
\label{app:sequence_mixer_details}

\begin{table}[!htbp]
\centering
\begin{threeparttable}
\caption{Recurrent state-update rules for linear-attention architectures.
The common readout is
$\mathbf{o}_t=\mathbf{S}_t^\top\mathbf{q}_t$.\tnote{k}}
\label{tab:linear-attention-updates}

\small
\setlength{\tabcolsep}{3pt}
\renewcommand{\arraystretch}{1.2}

\begin{tabularx}{\linewidth}{
    @{}
    l
    >{\centering\arraybackslash}X
    >{\raggedright\arraybackslash}p{2.25cm}
    @{}
}
\toprule
Architecture & State update & Mechanism \\
\midrule

LA\tnote{a}
&
$\displaystyle
\mathbf{S}_t
=
\mathbf{S}_{t-1}
+
\mathbf{k}_t\mathbf{v}_t^\top
$
&
Additive write
\\
RWKV-5\tnote{b}
&
$\displaystyle
\mathbf{S}_t
=
\operatorname{Diag}(\boldsymbol{\alpha})
\mathbf{S}_{t-1}
+
\mathbf{k}_t\mathbf{v}_t^\top
$
&
Fixed channel-wise decay
\\
Mamba-2\tnote{c}
&
$\displaystyle
\mathbf{S}_t
=
a_t\mathbf{S}_{t-1}
+
\mathbf{k}_t\mathbf{v}_t^\top
$
&
Scalar decay per head
\\
GLA\tnote{d}
&
$\displaystyle
\begin{aligned}
\mathbf{S}_t
={}&
\operatorname{Diag}(\boldsymbol{\alpha}_t)
\mathbf{S}_{t-1} \\
&+
\mathbf{k}_t\mathbf{v}_t^\top
\end{aligned}
$
&
Input-dependent channel-wise decay
\\
RWKV-6\tnote{e}
&
$\displaystyle
\mathbf{S}_t
=
\operatorname{Diag}(\boldsymbol{\alpha}_t)
\mathbf{S}_{t-1}
+
\mathbf{k}_t\mathbf{v}_t^\top
$
&
Input-dependent channel-wise decay
\\
DN\tnote{f}
&
$\displaystyle
\begin{aligned}
\mathbf{S}_t
={}&
\bigl(\mathbf{I}
-\beta_t\mathbf{k}_t\mathbf{k}_t^\top\bigr)
\mathbf{S}_{t-1}+\beta_t\mathbf{k}_t\mathbf{v}_t^\top
\end{aligned}
$
&
Delta update
\\

GDN\tnote{g}
&
$\displaystyle
\begin{aligned}
\mathbf{S}_t
={}&
\alpha_t
\bigl(\mathbf{I}
-\beta_t\mathbf{k}_t\mathbf{k}_t^\top\bigr)
\mathbf{S}_{t-1}
+
\beta_t\mathbf{k}_t\mathbf{v}_t^\top
\end{aligned}
$
&
Scalar decay and delta update
\\

KDA\tnote{h}
&
$\displaystyle
\begin{aligned}
\mathbf{S}_t
={}&
\bigl(\mathbf{I}
-\beta_t\mathbf{k}_t\mathbf{k}_t^\top\bigr)
\operatorname{Diag}(\boldsymbol{\alpha}_t)
\mathbf{S}_{t-1}
+
\beta_t\mathbf{k}_t\mathbf{v}_t^\top
\end{aligned}
$
&
Channel-wise decay and delta update
\\

GDN2\tnote{i}
&
$\displaystyle
\begin{aligned}
\mathbf{S}_t
={}&
\left(
\mathbf{I}
-\mathbf{k}_t
(\mathbf{b}_t\odot\mathbf{k}_t)^\top
\right)
\operatorname{Diag}(\boldsymbol{\alpha}_t)
\mathbf{S}_{t-1}
+
\mathbf{k}_t
(\mathbf{w}_t\odot\mathbf{v}_t)^\top
\end{aligned}
$
&
Channel-wise decay, erase, and write
\\
RWKV-7\tnote{j}
&
$\displaystyle
\begin{aligned}
\mathbf{S}_t
={}&
\bigl[
\operatorname{Diag}(\boldsymbol{\alpha}_t)
-
(\mathbf{a}_t\odot\hat{\mathbf{k}}_t)
\hat{\mathbf{k}}_t^\top
\bigr]\mathbf{S}_{t-1}
+
\mathbf{k}_t\mathbf{v}_t^\top
\end{aligned}
$
&
Channel-wise decay, rank-one erase and write
\\
\bottomrule
\end{tabularx}

\begin{tablenotes}[flushleft]
\footnotesize
\item[a] Linear Attention \citep{pmlr-v119-katharopoulos20a}.
\item[b] RWKV-5 (Eagle) \citep{peng2024eaglefinchrwkvmatrixvalued}.
\item[c] Mamba 2 \citep{mamba2}.
\item[d] Gated Linear Attention \citep{yang2024gatedlinearattentiontransformers}.
\item[e] RWKV-6 (Finch) \citep{peng2024eaglefinchrwkvmatrixvalued}.
\item[f] DeltaNet \citep{delta_net}.
\item[g] Gated DeltaNet \citep{yang2025gateddeltanetworksimproving}.
\item[h] Kimi Delta Attention \citep{kimiteam2025kimilinearexpressiveefficient}.
\item[i] Gated DeltaNet-2 \citep{hatamizadeh2026gateddeltanet2decouplingerase}.
\item[j] RWKV-7 \citep{peng2025rwkv7}.
\item[k] RWKV-5/6 and RWKV-7 slightly modify this readout with additional terms.
\end{tablenotes}

\end{threeparttable}
\end{table}

The table shows the update rule for a set of different SSMs, RNNs and linear attention models. Several of them overlap significantly with only minor differences separating how the update works. They do however deviate significantly from the standard transformer layout in their overall layer structure and are not explored in detail in this study.

The layer structure of linear transformer variants adds several additional components to the standard transformer layout, partly to account for the additional input dependences but also as they have been shown to improve performance generally. To isolate any differences between models to only the sequence mixer operation, we apply a standard setup following the convention used in Gated DeltaNet for all architectures across experiments. This allows for fair comparison between only the sequence mixers.

For an input $x_t=\bar x^{(\ell)}_t$, the mixers compute
\begin{align}
\tilde q_t &= W_qx_t,
\tilde k_t = W_kx_t,
\tilde v_t = W_vx_t, \beta_t = \sigma(W_\beta x_t )\\
\mathcal C_4(u)_t
    &= \sum_{\tau=0}^{3} c_{\tau}\odot u_{t-\tau},
\qquad u_s=0\ \text{for }s<1 \\
q_t &= \operatorname{Norm}(\operatorname{SiLU}\!\left(\mathcal C^{q}_4(\tilde q)_t\right)),\\
k_t &= \operatorname{Norm}(\operatorname{SiLU}\!\left(\mathcal C^{k}_4(\tilde k)_t\right)),\\
v_t &= \operatorname{SiLU}\!\left(\mathcal C^{v}_4(\tilde v)_t\right),\\
o_t &= \operatorname{SequenceMixer}(q_t,k_t,v_t) \\
\mathcal O_{\phi}(o_t,x_t)
   &=
   W_o\,
   \operatorname{vec}\!\left[
      \operatorname{RMSNorm}_{\varepsilon}(o_t)
      \odot \operatorname{SiLU}(W_gx_t)
   \right]
\end{align}

The delta-rule mixers use $\beta_t$ in their state updates as shown in the table; it is not a separate shared value gate. Additional mixer-specific projections, such as those controlling decay, are omitted here.
A particularly important component is the short convolution $\mathcal C_4$, a causal depthwise 1D convolution with a kernel width of four that is applied along the sequence dimension to capture local interactions between neighboring tokens. As in Gated DeltaNet, it is applied to the query, key, and value representations before the recurrent sequence-mixing operation. This local mixing is particularly relevant for associative recall since it allows for local token shifts, which provides a shortcut for forming induction heads in a single layer \citep{delta_net} enabling it to perform MQAR without additional layers.

The layers were implemented using the FLA library \citep{yang2024fla} with modifications to standardize layer parameters.

\FloatBarrier

\section{Model and technical details}
\label{app:llm_training_details}

The models consist of 20 layers with model dimension 1024, using 8 heads, yielding a total state size of 131,072 per layer. Each layer uses a sequence mixer interleaved with a SwiGLU \citep{swiglu} MLP with an intermediate dimension of 2752. The models are not parameter matched, as some require additional input projections for their update rules; for example, channel-wise decay requires more parameters than head-wise decay.

Each model is trained on 15B tokens, corresponding to approximately 44 tokens per parameter for a 340M-parameter model. Each step used a batch size of 128, with packed sequences of length 4,096 resulting in 524,288 tokens per step for 28,672 steps.

All models use a WSD \citep{wen2024understandingwarmupstabledecaylearningrates} schedule with 10\% warmup, 20\% cooldown, and a peak learning rate of $4.5\times10^{-4}$. The training framework was implemented with FLAME \citep{yang2025flame}.

We run evaluations using the \href{https://github.com/EleutherAI/lm-evaluation-harness}{lm-evaluation-harness} \citep{eval-harness} with minor modifications.

\FloatBarrier

\section{Data and Tokenizer Impact on Zoology analysis}
\label{app:zoology_data}

\begin{table}[!htbp]
\centering
\caption{Zoology-style AR-hit rate (\% of tokens) as a function of corpus and tokenizer.
A token is an AR-hit if it completes a bigram that appeared earlier in the context and occurs
$\leq$1,250 times per 10B tokens. Documents are chunked at 2048 tokens; bigram counts are taken
from held-out data of the same corpus and rescaled to 10B tokens. Pile = \texttt{pile-uncopyrighted}.}
\label{tab:ar_hit_corpus_tokenizer}
\begin{tabular}{llcc}
\toprule
Corpus & Tokenizer & AR-hits (\%) & Repeated bigrams (\%) \\
\midrule
\multirow{2}{*}{FineWeb-Edu} & Llama-2 (32k)   & 1.7 & 22.7 \\
                             & GPT-2 BPE (50k) & 2.4 & 19.9 \\
\midrule
\multirow{2}{*}{Pile}        & Llama-2 (32k)   & 3.8 & 40.8 \\
                             & GPT-2 BPE (50k) & 4.3 & 43.7 \\
\midrule
\multicolumn{2}{l}{Zoology \citet{arora2024zoology}, Pile} & 6.4 & -- \\
\bottomrule
\end{tabular}
\end{table}

Applying the Zoology AR-hit criterion yields fewer qualifying tokens than the reported 6.4\%. Both corpus composition and tokenization contribute to the difference (\autoref{tab:ar_hit_corpus_tokenizer}). The original study used the Pile \citep{gao2020pile800gbdatasetdiverse}; our comparison uses its uncopyrighted subset, which excludes Books3, BookCorpus2, OpenSubtitles, YouTube Subtitles, and OpenWebText2, together accounting for about 25\% of the original data.

With the Llama-2 tokenizer, repeated bigrams account for 40.8\% of tokens in this Pile subset versus 22.7\% in FineWeb-Edu, and AR-hit rates are 3.8\% versus 1.7\%. Using the GPT-2 BPE tokenizer \citep{radford2019language}, as in Zoology, raises the AR-hit rates to 4.3\% and 2.4\%, respectively. Corpus composition and the resulting frequency estimates may account for the remaining difference from Zoology's reported rate.

\FloatBarrier
\section{Recall prompts and context controls}
\label{app:example-cues}

The recall suite uses two prompt formats: the original benchmark
prompt and a prefix prompt consisting of the four words immediately
preceding the answer's first occurrence in the document.
For SQuAD, DROP, TriviaQA and NQ, the original prompt is a cloze statement;
for SWDE and FDA, it requests an attribute from the document.
The prefix replaces the original generation prompt and reproduces the local context in which the answer appeared. Both formats follow the supporting document.
\autoref{tab:recall-control-examples} illustrates these differences
and the answer substitutions used in the controls below.

\begin{table}[!htbp]
  \centering
  \footnotesize
  \setlength{\tabcolsep}{3pt}
  \renewcommand{\arraystretch}{1.1}
  \begin{tabularx}{\linewidth}{
    @{}l
    >{\raggedright\arraybackslash}X
    >{\raggedright\arraybackslash}p{0.24\linewidth}
    >{\raggedright\arraybackslash}p{0.15\linewidth}
    >{\raggedright\arraybackslash}p{0.15\linewidth}@{}
  }
    \toprule
    Dataset & Original prompt (end) & Four-word prefix
      & Answer & Substitute \\
    \midrule
    \multicolumn{5}{l}{\emph{Question answering}} \\
    DROP
      & ``The player who kicked the longest field goal is Shaun''
      & ``when the kicker Shaun''
      & Suisham
      & Disgust \\
    TQA
      & ``The singer who played the Goblin King Jareth in the film
         `Labyrinth' was''
      & ``Goblin King (played by''
      & David Bowie
      & Syrup Check \\
    NQ
      & ``The whistling in the Muppet movie was done by''
      & ``was performed by musician''
      & Andrew Bird
      & Matrix Cape \\
    \midrule
    \multicolumn{5}{l}{\emph{Document extraction}} \\
    SWDE
      & ``Summary of information above...\(\hookleftarrow\)director:''
      & ``for their survival. Director:''
      & Frank Marshall
      & Thank Homework \\
    FDA
      & ``Measurand:''
      & ``(AST) Systems. C. Measurand:''
      & Imipenem 0.25--16\,\(\mu\)g\/mL
      & Barstool 0.53--03\,\(\mu\)l\/xq \\
    FDA
      & ``Panel:''
      & ``Amplification System 4. Panel:''
      & Microbiology (83)
      & Anticipation (76) \\
    \bottomrule
  \end{tabularx}
  \caption{Example prompts and answer substitutions from the
  four-word context controls. Both prompt formats follow the
  supporting document in the full-context condition and stand
  alone when the document is removed.
  The original prompt uses the stock Based/JRT format.
  The prefix consists of the four words preceding the answer's
  first occurrence in the document.
  \(\hookleftarrow\) marks a newline.}
  \label{tab:recall-control-examples}
\end{table}

To examine how much performance depends on information in the
document, we evaluate the final 340M softmax and KDA models under
three conditions. In \emph{Full}, the document and prompt are
unchanged. In \emph{No ctx}, we remove the supporting document
and retain only the original generation prompt or four-word
prefix. In \emph{Swap}, every occurrence of the gold answer in the
supporting document is replaced with the same random substitute, matched in
character length and casing and within two tokens of the original
answer's token length. Alphabetic components are replaced with
dictionary words and digits with random digits.
We measure how often the output contains the substitute and
how often it contains the original answer.

All conditions use the same examples within each dataset,
restricted to documents where the answer appears verbatim,
the four-word cue is unique, and a valid replacement can be
constructed. This leaves 1,906 examples for DROP, 1,190 for
TriviaQA, 1,807 for NQ, 1,050 for SWDE, and 875 for FDA.
The resulting subsets differ from the main benchmark evaluation,
so the full-context scores below need not match its aggregate
results.

Generation is greedy, stops at a newline, and is limited to
48 generated tokens. Prompts are evaluated without truncation,
using a maximum length of 5120 tokens. Accuracy measures whether
the generated output contains the target answer; under
substitution, the target is the substitute.

\begin{table}[!htbp]
  \centering
  \small
  \setlength{\tabcolsep}{3pt}
  \begin{tabular}{llrrrrrrrr}
    \toprule
    & & \multicolumn{4}{c}{Softmax attention}
      & \multicolumn{4}{c}{KDA} \\
    \cmidrule(lr){3-6}
    \cmidrule(lr){7-10}
    & & & & \multicolumn{2}{c}{Swap}
      & & & \multicolumn{2}{c}{Swap} \\
    \cmidrule(lr){5-6}
    \cmidrule(lr){9-10}
    Dataset & Prompt
      & Full & No ctx & Sub. & Orig.
      & Full & No ctx & Sub. & Orig. \\
    \midrule
    \multirow{2}{*}{DROP}
      & Original & 19.7 & 2.4 & 7.0 & 4.6
                 & 22.0 & 1.6 & 6.6 & 6.7 \\
      & Prefix   & 67.8 & 5.1 & 54.1 & 5.3
                 & 57.5 & 5.8 & 38.8 & 4.9 \\
    \midrule
    \multirow{2}{*}{TQA}
      & Original & 44.0 & 13.6 & 16.4 & 13.4
                 & 42.7 & 15.1 & 14.4 & 14.6 \\
      & Prefix   & 83.6 & 7.2 & 73.8 & 3.8
                 & 69.8 & 7.1 & 60.3 & 4.8 \\
    \midrule
    \multirow{2}{*}{NQ}
      & Original & 23.4 & 5.1 & 9.0 & 3.9
                 & 15.3 & 5.3 & 2.7 & 5.6 \\
      & Prefix   & 49.4 & 1.2 & 40.1 & 1.4
                 & 19.1 & 1.5 & 12.1 & 2.3 \\
    \midrule
    \multirow{2}{*}{SWDE}
      & Original & 54.8 & 3.8 & 37.0 & 3.6
                 & 29.6 & 3.6 & 11.2 & 3.7 \\
      & Prefix   & 65.6 & 1.8 & 59.0 & 2.2
                 & 36.1 & 1.9 & 25.3 & 2.4 \\
    \midrule
    \multirow{2}{*}{FDA}
      & Original & 52.7 & 0.0 & 52.0 & 0.0
                 & 9.5 & 0.0 & 5.7 & 0.1 \\
      & Prefix   & 70.2 & 0.0 & 71.2 & 0.0
                 & 10.5 & 0.0 & 6.7 & 0.1 \\
    \bottomrule
  \end{tabular}
  \caption{Recall accuracy (\%) under context controls.
  \textbf{Full}: unchanged prompt with the supporting document.
  \textbf{No ctx}: supporting document removed.
  Under \textbf{Swap}, \textbf{Sub.} measures outputs containing
  the substituted answer and \textbf{Orig.} measures outputs
  containing the original answer.
  \textbf{Original} uses the benchmark's standard prompt;
  \textbf{Prefix} uses the four words preceding the answer's
  first occurrence in the document.
  All conditions use matched examples within each dataset.}
  \label{tab:recall-controls}
\end{table}

Removing the document substantially lowers accuracy for both
models under both prompt formats. With only the four-word
prefix, accuracy is at most 7.2\% and is closely matched between
softmax and KDA. The larger differences with the document present
therefore reflect differences in their use of the supplied
context. The original prompts also benefit substantially from
the document: on DROP and TriviaQA, the gains are comparable
between architectures, whereas NQ and the extraction tasks
already show a larger advantage for softmax.

Substitution further distinguishes the prompt formats.
For every dataset and both models, prefix prompts yield higher
substituted-answer accuracy than the original prompts.
For example, on TriviaQA, softmax reproduces the substitute
in 73.8\% of prefix-prompt examples compared to 16.4\% with
the original prompt; the corresponding KDA scores are
60.3\% and 14.4\%. Under prefix prompting, outputs containing
the original answer are uncommon, ranging from 0--5.3\%.
With the original TriviaQA prompt, however, both models
produce the original answer after substitution about as
often as they answer correctly without the document.

Substituted answers are generally harder to reproduce than
the original answers, even with a matching prefix. Since
the replacements need not be plausible answers to the
original question, this decrease combines sensitivity to
the document with difficulty generating an unusual or
incompatible answer. The substitution control therefore
supports the role of matching cues in eliciting retrieval,
while the context-removal results establish that both
prompt formats already make substantial use of the document.
\FloatBarrier
\section{MQAR Details}
\label{app:MQAR_details}

The synthetic MQAR task forms a sequence by sampling $N$ distinct keys without replacement, pairing each with a value, and placing the resulting pairs consecutively at the start of the sequence. The remainder of the sequence is filled with a designated noise token, and a subset of these positions is replaced by queries, each a repetition of one of the keys seen in the prefix. Writing keys as lowercase letters, values as digits, and the noise token as \texttt{0}, a
sequence takes the form
\begin{verbatim}
input:   a 3 b 7 c 1 0 0 0 b 0 0 a 0 0 c 0
target:  - - - - - - - - - 7 - - 3 - - 1 -
\end{verbatim}
\noindent
where \texttt{-} denotes a masked position. The name multi-query refers to the several lookups occurring within a single sequence, which prevents the model from succeeding by retaining only a single retrieved item. Difficulty is
controlled by the sequence length, the number of key--value pairs and the
vocabulary size.
\FloatBarrier
\section{Decay gates in large-scale copying}
\label{app:copy-gate-probe}

To investigate the lower copying accuracy of the diagonal-gated models, we inspect the final single-layer checkpoints at model
dimension 1024. All models have eight $128\times128$ heads and 131,072 state entries. The state is frozen after the copy
marker, so decay acts only while encoding the source.
We evaluate without retraining on $2^{20}$ copied tokens
per length.

H-GLA and GDN use one decay factor per head, whereas D-GLA
and KDA use a separate factor for each key channel.
We summarize retention by the channel half-life
\[
  \tau_{1/2}
  = \frac{\log(1/2)}
         {\frac{1}{c}\sum_{t=1}^{c}\log\alpha_t},
\]
where $\alpha_t$ is the retention factor and $c$ is the
source length. This measures decay independently of
the delta update.

The diagonal gates separate into two sharply distinct groups.
In D-GLA, 30.7\% of key channels have a half-life below one
token, increasing to 43.7\% in KDA. Within these groups,
median per-step retention is approximately $7\times10^{-6}$
and $10^{-8}$, respectively. Almost all remaining channels
retain at least 95.8\% over 4096 updates, with fewer than
0.4\% between these regimes. Rapidly decaying channels occur
in every head, leaving approximately 89 of 128 channels
per head in D-GLA and 72 in KDA with long retention times.

The scalar gates instead learn long retention times.
GDN's half-lives exceed $10^9$ tokens, while H-GLA's are
approximately $2\times10^5$, apart from one head near $10^4$.
Setting their decay to exactly one leaves accuracy unchanged
to three decimal places. The same intervention reduces
accuracy in the diagonal-gated models
(\autoref{tab:copy-gate-probe}). This contrast reflects their failure to learn near-unit retention, fixing decay to one leaves the head-wise models' learned dynamics effectively unchanged, but alters the dynamics on which D-GLA and KDA rely.

\begin{table}[!htbp]
  \centering
  \small
  \setlength{\tabcolsep}{5pt}
  \begin{tabular}{lrrr}
    \toprule
    & Channels with
    & \multicolumn{2}{c}{Copy accuracy (\%)} \\
    \cmidrule(lr){3-4}
    Model & $\tau_{1/2}<1$ (\%)
          & Learned decay & Decay fixed to 1 \\
    \midrule
    H-GLA & 0.0  & 83.7 & 83.7 \\
    GDN   & 0.0  & 88.0 & 88.0 \\
    D-GLA & 30.7 & 66.8 & 58.4 \\
    KDA   & 43.7 & 48.5 & 0.4 \\
    \bottomrule
  \end{tabular}
  \caption{Gate statistics and teacher-forced copying at
  length 4096. Decay is fixed to one during source encoding
  without retraining. Single-seed checkpoints.}
  \label{tab:copy-gate-probe}
\end{table}

The positional results distinguish this channel suppression
from the recency preference associated with the delta rule.
D-GLA's accuracy is nearly uniform across the source,
at 65.9\% in the first eighth and 67.7\% in the last.
KDA favors recent items (44.9\% versus 52.3\%), as do
DN (87.6\% versus 92.9\%) and GDN (85.3\% versus 90.6\%),
despite their absent or negligible decay. This shared
pattern is consistent with the delta-rule recency bias
observed in MQAR.

LA and DN achieve 91.4\% and 90.4\% accuracy with the same
state size. The diagonal-gated models therefore fall short
of the storage performance available at this scale, while
rapidly erasing information in a substantial fraction of
their channels. The deficit thus reflects failure to learn long retention in those channels, rather than a smaller nominal state. 

\FloatBarrier
\clearpage
\section{Complete LM Eval Results}
\label{app:Complete_Recall_Suite_Results}
\begin{table}[!htbp]
  \centering
  \footnotesize
  \setlength{\tabcolsep}{3pt}
  \renewcommand{\arraystretch}{1.0}
  \begin{tabular}{@{}llrrrrrrrr@{}}
    \toprule
    & & \multicolumn{2}{c}{Document extraction} & \multicolumn{4}{c}{In-context QA} & \multicolumn{2}{c}{Language modeling} \\
    \cmidrule(lr){3-4}\cmidrule(lr){5-8}\cmidrule(lr){9-10}
    Model & Prompt & SWDE & FDA & SQuAD & TQA & NQ & DROP & LMB & Wiki \\
    \midrule
    \multicolumn{10}{@{}l}{\textit{Standard}} \\
    Softmax & Base & 53.7 & 50.5 & 28.4 & 50.2 & 19.4 & 19.1 & 35.1 & 26.2 \\
    & Prefix & 65.8 & 69.5 & 74.0 & 84.0 & 48.9 & 67.7 & -- & -- \\
    \addlinespace[1pt]
    DN & Base & 21.1 & 8.2 & 29.3 & 44.8 & 12.4 & 19.1 & 37.8 & 27.9 \\
    & Prefix & 26.6 & 11.9 & 58.5 & 70.4 & 17.5 & 51.3 & -- & -- \\
    \addlinespace[1pt]
    GLA & Base & 20.0 & 4.4 & 31.4 & 47.5 & 13.4 & 18.0 & 30.5 & 27.7 \\
    & Prefix & 20.9 & 5.1 & 51.5 & 65.0 & 14.7 & 50.4 & -- & -- \\
    \addlinespace[1pt]
    GDN & Base & 21.9 & 7.9 & 31.9 & 46.1 & 14.2 & 17.8 & 33.1 & 27.2 \\
    & Prefix & 31.7 & 10.0 & 59.5 & 70.0 & 19.2 & 50.4 & -- & -- \\
    \addlinespace[1pt]
    KDA & Base & 29.4 & 12.6 & 34.8 & 49.6 & 14.8 & 21.6 & 29.7 & 25.9 \\
    & Prefix & 36.7 & 14.3 & 64.3 & 70.3 & 18.8 & 57.1 & -- & -- \\
    \midrule
    \multicolumn{10}{@{}l}{\textit{Sentence Repetition}} \\
    Softmax & Base & 53.6 & 62.7 & 25.6 & 50.4 & 19.3 & 18.0 & 36.4 & 26.3 \\
    & Prefix & 71.7 & 84.2 & 81.3 & 87.6 & 63.7 & 80.2 & -- & -- \\
    \addlinespace[1pt]
    DN & Base & 27.1 & 22.6 & 31.7 & 45.2 & 12.6 & 17.8 & 38.6 & 28.8 \\
    & Prefix & 45.4 & 37.3 & 72.3 & 78.8 & 30.0 & 67.1 & -- & -- \\
    \addlinespace[1pt]
    GLA & Base & 28.7 & 18.9 & 33.6 & 48.2 & 15.2 & 18.1 & 32.2 & 28.5 \\
    & Prefix & 38.1 & 26.1 & 69.1 & 75.5 & 25.2 & 63.4 & -- & -- \\
    \addlinespace[1pt]
    GDN & Base & 30.0 & 23.5 & 33.0 & 46.2 & 15.1 & 18.3 & 33.4 & 28.0 \\
    & Prefix & 47.8 & 41.2 & 75.3 & 80.1 & 30.9 & 69.1 & -- & -- \\
    \addlinespace[1pt]
    KDA & Base & 34.7 & 27.4 & 37.3 & 48.9 & 16.5 & 20.1 & 29.8 & 26.5 \\
    & Prefix & 52.0 & 44.3 & 75.9 & 78.2 & 32.2 & 69.8 & -- & -- \\
    \midrule
    \multicolumn{10}{@{}l}{\textit{TLCE}} \\
    Softmax & Base & 61.7 & 59.6 & 29.2 & 56.2 & 26.8 & 21.3 & 35.5 & 27.7 \\
    & Prefix & 71.8 & 69.9 & 77.4 & 85.2 & 62.1 & 72.0 & -- & -- \\
    \addlinespace[1pt]
    DN & Base & 31.8 & 10.9 & 33.5 & 49.9 & 18.3 & 20.4 & 38.9 & 29.5 \\
    & Prefix & 41.1 & 13.0 & 63.5 & 75.4 & 28.2 & 58.4 & -- & -- \\
    \addlinespace[1pt]
    GLA & Base & 31.6 & 8.7 & 35.6 & 53.8 & 19.0 & 19.1 & 31.5 & 29.1 \\
    & Prefix & 35.8 & 9.5 & 55.7 & 70.5 & 24.7 & 56.5 & -- & -- \\
    \addlinespace[1pt]
    GDN & Base & 33.8 & 13.1 & 37.0 & 52.7 & 19.8 & 20.1 & 33.7 & 28.7 \\
    & Prefix & 46.7 & 16.3 & 61.5 & 75.0 & 29.5 & 56.6 & -- & -- \\
    \addlinespace[1pt]
    KDA & Base & 41.7 & 17.2 & 41.9 & 57.0 & 19.6 & 22.3 & 30.9 & 27.4 \\
    & Prefix & 50.3 & 21.1 & 69.1 & 76.8 & 30.7 & 62.2 & -- & -- \\
    \midrule
    \multicolumn{10}{@{}l}{\textit{TLCE + Repeat}} \\
    Softmax & Base & 60.6 & 70.2 & 37.7 & 56.5 & 28.9 & 21.4 & 37.7 & 27.9 \\
    & Prefix & 74.8 & 86.2 & 82.5 & 89.2 & 74.7 & 81.6 & -- & -- \\
    \addlinespace[1pt]
    DN & Base & 37.5 & 30.4 & 36.5 & 51.3 & 18.3 & 20.2 & 38.6 & 30.9 \\
    & Prefix & 54.1 & 50.1 & 74.3 & 83.9 & 42.5 & 69.3 & -- & -- \\
    \addlinespace[1pt]
    GLA & Base & 34.2 & 24.2 & 37.4 & 53.7 & 20.7 & 19.3 & 33.3 & 30.5 \\
    & Prefix & 44.4 & 43.2 & 70.0 & 80.8 & 38.5 & 68.2 & -- & -- \\
    \addlinespace[1pt]
    GDN & Base & 36.4 & 32.8 & 38.2 & 51.3 & 20.5 & 21.6 & 32.7 & 29.9 \\
    & Prefix & 55.2 & 52.4 & 74.4 & 83.6 & 44.6 & 70.2 & -- & -- \\
    \addlinespace[1pt]
    KDA & Base & 40.3 & 37.9 & 42.6 & 56.1 & 20.8 & 20.9 & 31.0 & 28.5 \\
    & Prefix & 56.2 & 58.2 & 76.9 & 83.0 & 47.8 & 74.5 & -- & -- \\
    \midrule
    \multicolumn{2}{@{}l}{Sequence length} & 1{,}302 & 2{,}271 & 306 & 346 & 1{,}832 & 309 & 87 & 5{,}195 \\
    \bottomrule
  \end{tabular}
  \caption{LM Evaluation results for each training variant: Standard, Sentence Repetition, TLCE, and TLCE combined with Repeat. Base uses the original evaluation prompt; Prefix replaces it with the four-word prefix. Higher accuracy is better for document extraction and QA; lower word perplexity is better for language modeling. ``--'' indicates results not evaluated.}
    \label{tab:appendix-eval}
\end{table}
\FloatBarrier

\begin{table}[htbp]
\centering
\begingroup
\small
\setlength{\tabcolsep}{3pt}
\resizebox{\linewidth}{!}{%
\begin{tabular}{@{}lrrrrrr@{}}
\toprule
& & \multicolumn{4}{c}{Recall accuracy (\%)} & LM \\
\cmidrule(lr){3-6}
Method & Loss $\downarrow$ & All $\uparrow$ & QA $\uparrow$ &
Extr. $\uparrow$ & Prefix $\uparrow$ & PPL $\downarrow$ \\
\midrule
Baseline
& 2.3439
& 27.1
& 30.2
& 21.0
& 43.6
& 27.84 \\
\midrule
\multicolumn{7}{l}{\textit{Repetition}} \\
Sentence 5\%
& 2.3510 (+0.0071)
& 30.0 (+2.9)
& 30.2 (+0.0)
& 29.5 (+8.5)
& 56.2 (+12.6)
& 27.95 (+0.11) \\
Sentence 10\%
& 2.3562 (+0.0123)
& 30.8 (+3.7)
& 30.7 (+0.5)
& 31.1 (+10.1)
& 58.7 (+15.2)
& 28.15 (+0.32) \\
Sentence 22\%
& 2.3680 (+0.0241)
& 30.8 (+3.7)
& 30.2 (+0.0)
& 32.1 (+11.1)
& 61.2 (+17.6)
& 29.12 (+1.29) \\
Sentence 58\%
& 2.4126 (+0.0686)
& 29.1 (+2.0)
& 28.9 (-1.3)
& 29.7 (+8.7)
& 60.6 (+17.0)
& 30.78 (+2.95) \\
Sequence Repetition 10\%
& 2.3544 (+0.0105)
& 30.0 (+2.9)
& 30.8 (+0.6)
& 28.4 (+7.4)
& 55.9 (+12.3)
& 28.09 (+0.25) \\
Sequence Repetition 12.5\%
& 2.3565 (+0.0126)
& 29.8 (+2.7)
& 30.3 (+0.1)
& 28.6 (+7.6)
& 56.8 (+13.2)
& 28.28 (+0.45) \\
Sequence Repetition 25\%
& 2.3665 (+0.0226)
& 29.7 (+2.6)
& 29.8 (-0.4)
& 29.5 (+8.5)
& 57.4 (+13.8)
& 29.05 (+1.22) \\
\midrule
\multicolumn{7}{l}{\textit{Loss reweighting}} \\
AR-hit $\times5$
& 2.3602 (+0.0163)
& 29.0 (+1.9)
& 31.8 (+1.6)
& 23.6 (+2.6)
& 49.5 (+5.9)
& \textbf{27.78 (-0.06)} \\
LongCE
& 2.4306 (+0.0867)
& 32.9 (+5.8)
& 34.6 (+4.4)
& 29.3 (+8.3)
& 52.3 (+8.7)
& 31.57 (+3.73) \\
TLCE
& 2.3772 (+0.0333)
& 33.3 (+6.2)
& \textbf{35.2 (+5.0)}
& 29.5 (+8.5)
& 51.7 (+8.1)
& 29.15 (+1.32) \\
\midrule
\multicolumn{7}{l}{\textit{Combined}} \\
TLCE + Sentence 10\%
& 2.3977 (+0.0538)
& \textbf{36.4 (+9.3)}
& 35.1 (+4.9)
& \textbf{39.1 (+18.1)}
& \textbf{66.1 (+22.5)}
& 29.71 (+1.88) \\
\bottomrule
\end{tabular}%
}
\endgroup
\caption{Selected KDA 340M cooldown ablations. Parentheses show
changes from baseline; accuracy changes are percentage points.
Bold marks the largest improvement over baseline in each benchmark
column; no intervention improves held-out loss. Loss is held-out
cross-entropy in nats. All averages the six original recall tasks;
QA averages DROP, TriviaQA, NQ and SQuAD; Extr. averages SWDE and
FDA. Prefix averages all six tasks with four-word prefix prompts.
Baseline, Sentence 10\%, TLCE, and TLCE + Sentence 10\% recall
averages are calculated from the displayed task scores in
\autoref{tab:appendix-eval}; their changes are calculated before
rounding. For other settings, changes are calculated from the
displayed ablation scores and may differ by 0.1 from changes
calculated using unrounded scores. PPL is the arithmetic mean of
WikiText and LAMBADA perplexities, not a pooled perplexity.
Sentence copies sentence spans; Sequence repeats full sequences.
For Sentence Repetition, percentages specify the targeted
copied-token share; for Sequence Repetition, they specify the
probability of appending a full copy. These targets need not equal
realized copied-token shares. All reweighting variants normalize
by total weight.}
\label{tab:recall-ablation-summary}
\end{table}

\section{Recall Training Ablations}
\label{app:recall-training-ablations}

\subsection{Cooldown Ablations at 340M}
\autoref{tab:recall-ablation-summary} compares interventions applied during the final 20\% of KDA 340M training, with the starting checkpoint, data order, schedule and step budget held fixed. Each arm uses one training seed, so small differences should not be interpreted as robust rankings.

\paragraph{Repetition}
Sequence repetition appends a full copy after a selected training sequence, forming $[x;x]$; its rate is the probability of repeating a sequence. Sentence repetition instead copies shorter spans, averaging 64 tokens, from sentence boundaries and inserts them at randomly selected later sentence boundaries within the same context. Its rate specifies the targeted fraction of copied tokens. Both retain causal next-token supervision, but sentence repetition intersperses copying with ordinary language modeling and exposes the model to varied retrieval distances and shorter matching prefixes.

Sentence repetition improves extraction and prefix recall, with diminishing returns as its share increases. Raising the nominal sentence rate from 10\% to 22\% adds only 1.0 extraction points and 2.5 prefix points, while nearly doubling the held-out loss increase from 0.0123 to 0.0241 nats. At 58\%, recall falls relative to 22\% and language modeling loss increases further. Sequence Repetition 12.5\% gives smaller extraction and prefix scores than Sentence 10\% (28.6\% versus 31.1\%, and 56.8\% versus 58.7\%) at comparable held-out loss cost (0.0126 versus 0.0123 nats). Increasing the sequence rate to 25\% still trails Sentence 10\% on both measures, while increasing the loss cost to 0.0226 nats. These results support the 10\% sentence setting as a compromise between recall and language modeling.

\paragraph{Loss reweighting}
We first evaluate AR-hit $\times5$, assigning weight five to tokens completing a rare bigram that appeared earlier in the context and weight one otherwise. This targets approximately $1.7\%$ of tokens and requires no additional model forward pass. Although it substantially reduces loss on AR-hit tokens, its gains on the recall suite are smaller than those of either LongCE variant. QA and extraction accuracy reach 31.8\% and 23.6\%, compared with 34.6\% and 29.3\% for unthresholded LCE. Its language modeling cost is lower, however, and LAMBADA accuracy improves from 34.0\% to 35.3\%.

This limited transfer is consistent with \citet{fang2025wrongperplexitylongcontextlanguage}, who find that perplexity on repeated $n$-grams correlates much less strongly with long-context benchmark performance than their context-sensitive LongPPL metric. We therefore evaluate LongCE, which uses the detached long--short log-probability difference as
\begin{equation}
\Delta_i=\operatorname{sg}\!\left[\log p_{\mathrm{full}}(y_i)-\log p_{\mathrm{short}}(y_i)\right].
\end{equation}
For eligible tokens, the two weighting rules are
\begin{equation}
w_i^{\mathrm{LongCE}}=\min\{e^{\Delta_i},5\},
\qquad
w_i^{\mathrm{TLCE}}=
\begin{cases}
\min\{e^{\Delta_i},5\}, & \Delta_i>1,\\
1, & \Delta_i\leq1.
\end{cases}
\end{equation}
Eligibility requires at least 128 tokens since the preceding reset in the additional, gradient-free pass with 512-token windows; other valid tokens receive weight one in both variants. Both experimental variants use the same normalized objective,
\begin{equation}
\mathcal L=-\frac{\sum_i w_i\log p_{\mathrm{full}}(y_i)}{\sum_i w_i},
\end{equation}
with sums over valid targets and no gradients through the weights. We normalize by the total weight to keep the mean effective token weight at one, avoiding changes in gradient scale caused solely by the average weight and allowing us to retain the baseline learning-rate schedule. LCE assigns raw weights below one when $\Delta_i<0$ and above one when $\Delta_i>0$. TLCE instead leaves all raw weights at one unless the context benefit exceeds one nat. It therefore preserves the relative weighting among unselected tokens while retaining LCE's capped weighting for large positive differences. Normalization still reduces unselected tokens' effective weights when other tokens are upweighted, however it does reduce the impact of very weakly improved tokens, as these are unlikely to be real recall hits. The shorter scoring windows may make this threshold useful, as many tokens may benefit slightly from context beyond 128 tokens without requiring precise retrieval.

TLCE retains recall performance close to unthresholded LongCE while reducing the held-out loss increase from 0.0867 to 0.0333 nats. Both use the same weight normalization, isolating the threshold-and-unit-weight rule.

\paragraph{Complementary effects}
Sentence repetition and TLCE strengthen recall incentives in different ways, by introducing repeated text and upweighting tokens whose prediction benefits from longer context, respectively. When combined, TLCE can also upweight repeated tokens when their prediction benefits sufficiently from the earlier occurrence. The interventions nevertheless show distinct strengths. Sentence repetition at 10\% raises FDA accuracy from 12.6\% to 27.4\%, compared with 17.2\% for TLCE, whereas TLCE improves SWDE from 29.4\% to 41.7\%, compared with 34.7\% for repetition. Combining them yields 37.9\% on FDA and 40.3\% on SWDE, for a mean extraction accuracy of 39.1\%, exceeding that of either intervention alone. It also raises prefix accuracy to 66.1\% while retaining the QA performance of TLCE alone (35.1\% versus 35.2\%). These complementary gains come at a larger held-out loss increase of 0.0538 nats.

\subsection{Intervention sweeps at 60M}
\label{app:recall-training-ablations-60m}

We extend the intervention sweep using 61.7M-parameter KDA models
trained on 2.52B FineWeb-Edu tokens at context length 4096.
Cooldown interventions restart the same checkpoint and use the same
data order and optimization schedule for the final 10\% of training.
\autoref{tab:recall-ablation-60m} reports selected settings spanning
the repetition range and alternative loss weightings.
We separate document extraction from QA, as in the 340M experiments,
and additionally report answer loss because exact-match accuracy is
low on several tasks at this scale. Each setting uses one training
seed.

\begin{table}[htbp]
\centering
\small
\setlength{\tabcolsep}{4pt}
\begin{tabular}{@{}lrrrrrr@{}}
\toprule
& LM & \multicolumn{3}{c}{Original-prompt answer loss}
& \multicolumn{2}{c}{Prefix} \\
\cmidrule(lr){3-5}\cmidrule(lr){6-7}
Method & Loss $\downarrow$
& SWDE $\downarrow$ & FDA $\downarrow$ & QA $\downarrow$
& Loss $\downarrow$ & Acc.\ $\uparrow$ \\
\midrule
Baseline & 2.963 & 2.739 & 1.859 & 2.048 & 1.908 & 16.1 \\
\midrule
\multicolumn{7}{@{}l}{\textit{Sentence repetition: changes from baseline}} \\
Sentence 5\%  & +0.017 & -0.143 & -0.351 & -0.033 & -0.284 & +4.6 \\
Sentence 9\% (short)
              & +0.027 & -0.177 & -0.402 & -0.034 & -0.329 & +5.9 \\
Sentence 9\%  & +0.029 & -0.254 & -0.563 & -0.045 & -0.453 & +8.0 \\
Sentence 20\% & +0.057 & -0.296 & -0.732 & -0.042 & -0.592 & +11.3 \\
Sentence 37\% & +0.105 & -0.369 & -0.783 & -0.012 & -0.673 & +14.5 \\
Sentence 44\% & +0.130 & -0.306 & -0.737 & +0.029 & -0.635 & +13.5 \\
\midrule
\multicolumn{7}{@{}l}{\textit{Sequence repetition: changes from baseline}} \\
Sequence 8\%  & +0.021 & -0.287 & -0.374 & -0.017 & -0.309 & +4.6 \\
Sequence 18\% & +0.045 & -0.381 & -0.529 & -0.028 & -0.462 & +7.3 \\
Sequence 50\% & +0.128 & -0.394 & -0.658 & +0.011 & -0.640 & +13.2 \\
\midrule
\multicolumn{7}{@{}l}{\textit{Loss reweighting: changes from baseline}} \\
AR-hit $\times5$
              & +0.022 & -0.161 & -0.132 & -0.096 & -0.203 & +3.9 \\
AR-hit $\times6.5$
              & +0.032 & -0.179 & -0.141 & -0.104 & -- & -- \\
LongCE        & +0.076 & -0.653 & -0.389 & -0.218 & -0.517 & +8.4 \\
TLCE          & +0.032 & -0.473 & -0.334 & -0.169 & -0.390 & +6.3 \\
\bottomrule
\end{tabular}
\caption{Selected KDA 60M cooldown ablations. Baseline entries are
absolute; all other entries are changes from baseline.
Losses are in nats and accuracy changes are percentage points.
LM is held-out FineWeb-Edu loss, evaluated on 64 documents of
1024 tokens. QA averages DROP, TriviaQA and NQ; Prefix averages
SWDE, FDA and these three QA tasks with four-word prefix prompts.
Recall uses 400 examples per task and teacher-forced evaluation;
exact match requires every answer token to be the argmax.
Repetition percentages are rounded measured shares of scored
copied tokens, unlike the nominal settings in
\autoref{tab:recall-ablation-summary}.
Standard sentence spans average 64 tokens; ``short'' uses
8--32-token spans. Dashes indicate unavailable evaluations.}
\label{tab:recall-ablation-60m}
\end{table}

\paragraph{Repetition strength and span length}
The smaller models reproduce the recall--language modeling trade-off
seen at 340M. Sentence repetition near 10\% reduces prefix answer
loss by 0.453 nats and raises prefix accuracy by 8.0 points, at a
held-out loss cost of 0.029 nats. Increasing the share to 20\%
approximately doubles this cost for a further 3.3 prefix points.
Larger shares continue to improve prefix recall up to the 37\%
setting shown, but provide little additional QA benefit; at 44\%,
both prefix loss and QA loss worsen relative to lower rates.
At comparable repetition shares, shorter spans yield smaller
prefix-loss reductions (0.329 versus 0.453 nats).
Sentence repetition also achieves a prefix-loss reduction close
to Sequence 18\% at lower LM cost (0.029 versus 0.045 nats).
These results support sentence repetition near 10\% as a
low-cost compromise rather than a recall-maximizing setting.

\paragraph{Complementary effects of loss reweighting}
TLCE improves SWDE and QA more than sentence repetition near 10\%,
whereas repetition produces larger FDA and prefix gains at a
similar LM cost. Relative to LongCE, TLCE retains approximately
three quarters of the mean QA and prefix loss reductions while
reducing the LM penalty from 0.076 to 0.032 nats.
AR-hit reweighting transfers less strongly: even at $\times6.5$,
which approximately matches TLCE's additional weight share and
has the same measured LM cost, its SWDE, FDA and QA improvements
are smaller. Together, these findings support the complementary
roles of sentence repetition and TLCE observed at 340M.

\paragraph{Extending exposure}
Applying sentence repetition throughout training substantially improves
recall relative to introducing it only during the final 10\%, with
little change in the measured language modeling cost
(\autoref{tab:repeat-exposure-60m}).
Near a 10\% repetition share, the prefix accuracy gain increases from
8.0 to 21.4 percentage points, while the held-out LM loss penalty
remains at 0.029 nats.
Near 20\%, the gain increases from 11.3 to 25.9 points, while the
LM penalty changes from 0.057 to 0.060 nats.
Prefix answer-loss reductions also become substantially larger.
Thus, longer exposure more than doubles the prefix accuracy gains
at comparable LM cost, despite replacing fresh text over a much
larger fraction of training.

\begin{table}[htbp]
\centering
\small
\setlength{\tabcolsep}{5pt}
\begin{tabular}{@{}llrrr@{}}
\toprule
Repeat share & Exposure
& $\Delta$ LM loss $\downarrow$
& $\Delta$ Prefix loss $\downarrow$
& $\Delta$ Prefix acc.\ $\uparrow$ \\
\midrule
9.3\%  & Cooldown   & +0.029 & -0.453 & +8.0 \\
10.0\% & Throughout & +0.029 & -0.909 & +21.4 \\
\midrule
19.8\% & Cooldown   & +0.057 & -0.592 & +11.3 \\
20.0\% & Throughout & +0.060 & -1.066 & +25.9 \\
\bottomrule
\end{tabular}
\caption{Sentence-repetition exposure at 61.7M parameters.
All entries are changes from the standard-training baseline.
Losses are in nats and accuracy changes are percentage points.
Prefix averages the five exact-prefix tasks used in
\autoref{tab:recall-ablation-60m}.
Cooldown applies repetition during the final 10\% of training;
Throughout applies it from the first step.
Shares are measured copied-token shares and are approximately matched.
All runs use the same total training budget; each setting uses one seed.}
\label{tab:repeat-exposure-60m}
\end{table}

The full-training KDA experiment at 340M provides complementary
evidence with both sentence repetition and TLCE enabled.
Relative to applying both only during cooldown, document extraction
accuracy rises from 39.1\% to 48.9\%, approaching standard-trained
softmax at 52.1\%, while the mean LAMBADA/WikiText perplexity penalty
is comparable at 1.8 versus 1.9.
Although these comparisons do not isolate the contribution of
fresh-content replacement, they show that longer exposure can
substantially improve recall without a correspondingly larger
measured LM penalty.

\paragraph{Position dependence}
Copying improvements are not uniform across context positions.
For Sentence 20\%, a random-span probe at a retrieval distance of
512 tokens improves from 18\% to 45\% accuracy near position 1024,
but only from 10\% to 18\% near position 4080.
Aligning both copies with sentence boundaries raises the latter
to 31\%. These diagnostics motivate distributing repeated targets
throughout the context and checking transfer across positions
and span boundaries.

\FloatBarrier
\clearpage
\section{AR-hits under recall training}
\label{app:AR-hits-repeat}

\begin{table}[htbp]
  \centering
  \small
  \setlength{\tabcolsep}{3pt}
  \begin{tabular}{l r r r r r r}
    \toprule
    Group & Count & Softmax & DN & GLA & GDN & KDA \\
    \midrule
    \multicolumn{7}{l}{\emph{Standard run}} \\
    \quad AR-hits, 1--511 back & 226,922 & 0.770$^*$ & $+0.122$ & $+0.153$ & $+0.094$ & $+0.072$ \\
    \quad AR-hits, 512+ back & 46,148 & 1.454$^*$ & $+0.462$ & $+0.559$ & $+0.391$ & $+0.334$ \\
    \multicolumn{7}{l}{\emph{Rare tokens}} \\
    \quad New prefix, 1--511 back & 16,158 & 4.027$^*$ & $+0.135$ & $+0.071$ & $+0.042$ & $-0.106$ \\
    \quad New prefix, 512+ back & 3,541 & 5.355$^*$ & $+0.148$ & $+0.197$ & $+0.135$ & $+0.019$ \\
    \quad Same prefix, 1--511 back & 12,472 & 1.794$^*$ & $+0.219$ & $+0.235$ & $+0.176$ & $+0.127$ \\
    \quad Same prefix, 512+ back & 3,158 & 2.409$^*$ & $+0.657$ & $+0.850$ & $+0.628$ & $+0.521$ \\
    \midrule
    \multicolumn{7}{l}{\emph{Repeat run}} \\
    \quad AR-hits, 1--511 back & 226,922 & $-0.002$ & $+0.132$ & $+0.153$ & $+0.092$ & $+0.074$ \\
    \quad AR-hits, 512+ back & 46,148 & $-0.022$ & $+0.352$ & $+0.372$ & $+0.295$ & $+0.229$ \\
    \multicolumn{7}{l}{\emph{Rare tokens}} \\
    \quad New prefix, 1--511 back & 16,158 & $+0.016$ & $+0.190$ & $+0.145$ & $+0.107$ & $-0.067$ \\
    \quad New prefix, 512+ back & 3,541 & $+0.028$ & $+0.204$ & $+0.254$ & $+0.200$ & $+0.059$ \\
    \quad Same prefix, 1--511 back & 12,472 & $-0.006$ & $+0.230$ & $+0.235$ & $+0.176$ & $+0.146$ \\
    \quad Same prefix, 512+ back & 3,158 & $-0.031$ & $+0.463$ & $+0.551$ & $+0.411$ & $+0.342$ \\
    \midrule
    \multicolumn{7}{l}{\emph{TLCE run}} \\
    \quad AR-hits, 1--511 back & 226,922 & $-0.014$ & $+0.111$ & $+0.145$ & $+0.074$ & $+0.060$ \\
    \quad AR-hits, 512+ back & 46,148 & $-0.351$ & $+0.012$ & $+0.076$ & $-0.047$ & $-0.096$ \\
    \multicolumn{7}{l}{\emph{Rare tokens}} \\
    \quad New prefix, 1--511 back & 16,158 & $-0.100$ & $+0.038$ & $+0.006$ & $-0.040$ & $-0.176$ \\
    \quad New prefix, 512+ back & 3,541 & $-0.602$ & $-0.419$ & $-0.405$ & $-0.456$ & $-0.620$ \\
    \quad Same prefix, 1--511 back & 12,472 & $-0.029$ & $+0.182$ & $+0.211$ & $+0.131$ & $+0.095$ \\
    \quad Same prefix, 512+ back & 3,158 & $-0.492$ & $+0.015$ & $+0.115$ & $-0.029$ & $-0.153$ \\
    \midrule
    \multicolumn{7}{l}{\emph{TLCE+Repeat run}} \\
    \quad AR-hits, 1--511 back & 226,922 & $-0.013$ & $+0.111$ & $+0.141$ & $+0.073$ & $+0.062$ \\
    \quad AR-hits, 512+ back & 46,148 & $-0.366$ & $-0.043$ & $-0.029$ & $-0.089$ & $-0.146$ \\
    \multicolumn{7}{l}{\emph{Rare tokens}} \\
    \quad New prefix, 1--511 back & 16,158 & $-0.063$ & $+0.097$ & $+0.111$ & $+0.024$ & $-0.132$ \\
    \quad New prefix, 512+ back & 3,541 & $-0.558$ & $-0.308$ & $-0.292$ & $-0.342$ & $-0.534$ \\
    \quad Same prefix, 1--511 back & 12,472 & $-0.022$ & $+0.187$ & $+0.217$ & $+0.139$ & $+0.115$ \\
    \quad Same prefix, 512+ back & 3,158 & $-0.509$ & $-0.084$ & $-0.042$ & $-0.125$ & $-0.193$ \\
    \bottomrule
  \end{tabular}
  \caption{Loss on both recall tasks for each mixer and training variant, held-out FineWeb-Edu, split at 512 tokens. Every entry is the loss in nats minus that of softmax attention with standard training, so all panels share one scale; the Softmax column gives the softmax models trained with each variant. $^*$The standard softmax model is the reference itself, so its entries are its absolute loss in nats instead of a difference. The gap of a linear mixer to the softmax model trained the same way is its entry minus the Softmax entry in the same row. TLCE+Repeat uses sentence-aligned splices.}
  \label{tab:gap-closed-512}
\end{table}

We compare all training variants against softmax attention with standard training (\autoref{tab:gap-closed-512}), assessing how closely the linear mixers approach this fixed baseline. Both Repeat and TLCE lower AR-hit loss mainly at recall distances of 512 tokens or more.

At these distances, Repeat lowers the loss of the linear mixers by 0.10--0.19 nats, although all four remain above the baseline. TLCE reduces their loss by 0.43--0.48 nats, bringing DN and GLA within 0.08 nats of the baseline and GDN and KDA below it. Combining TLCE and Repeat brings all four linear mixers below the baseline, by 0.03--0.15 nats on AR-hits and 0.04--0.19 nats on rare tokens with the same prefix. On rare tokens with a new prefix at these distances, both TLCE and TLCE+Repeat also bring all four mixers below the baseline.

At distances below 512 tokens, changes in AR-hit loss are small, and the linear mixers remain 0.06--0.15 nats above the baseline across variants. These comparisons establish that the training schemes can close the gap to standard-training softmax on the longer-distance recall tasks. Softmax also benefits from these schemes: TLCE and TLCE+Repeat lower its long-distance AR-hit loss by 0.35 and 0.37 nats, respectively, and it retains a lower loss than every linear mixer trained with the same variant on those tokens.

\FloatBarrier
\clearpage
\section{Detailed NIAH Results}
\label{app:NIAH_Results}

Tables~\ref{tab:niah-single} and~\ref{tab:niah-multi} report per-task NIAH accuracy at context lengths of 2K, 4K, and 8K for all evaluated models and training variants.

\paragraph{Prompt sensitivity and length}
The effects of training on NIAH generation depend strongly on the task, prompt, and context length. Repetition produces its strongest gains on word retrieval with exact-prefix prompts, consistent with closer alignment between the training objective and the evaluation cue. Transfer to other formats is less reliable. On standard-prompt NIAH-2 at 4K, repetition raises KDA accuracy from 55\% to 58\%, but lowers GDN from 48\% to 21\%, DN from 24\% to 11\%, and GLA from 30\% to 7\%. These declines indicate a limitation of repetition's benefits beyond closely aligned prompts.

Other aggregate changes are dominated by individual failures. At 4K, combined training lowers GDN's NIAH-2 accuracy from 72\% to 4\% under the colon prompt, outweighing improvements in the other three recurrent architectures. With the exact prefix, however, GDN improves from 62\% to 88\%. Teacher-forced accuracy changes less, from 95\% to 88\%. Per-model and per-prompt results are therefore necessary to distinguish widespread declines from localized failures.

Generation accuracy also varies non-monotonically with length under the same prompt. Full KDA scores 99\%, 40\%, and 97\% on standard-prompt NIAH-2 at 2K, 4K, and 8K, respectively, while teacher-forced token accuracy remains at 100\%, 95\%, and 100\%. This intermediate-length failure does not follow a simple deterioration with retrieval distance. TLCE also exhibits localized failures, despite improving word retrieval across prompt formats, so neither intervention provides uniformly reliable generation gains.

Teacher-forced results show more consistent aggregate improvements. At 4K, both interventions improve the recurrent-model average on every single-needle task, and both improve the multi-needle average under all evaluated formats. Longer exposure further improves performance. Full KDA retains 93\% mean token accuracy over NIAH-1--4 at 8K and retrieves 66\% of word values with the exact prefix. These gains coexist with substantial complete-answer generation failures. Teacher forcing measures token prediction given the correct preceding answer tokens, so improved token-level recall does not by itself establish reliable production of the full answer. 
\paragraph{Task differences}
The delta-rule models (KDA, GDN, DN) reach at least 87\% on NIAH-1 at every length and prompt, whereas baseline GLA falls from 10\% at 2K to 0\% at 8K under the standard prompt. Because the delta rule overwrites associations bound to a repeated key, the repeated haystack sentence is largely ignored and the needle is preserved. On the multi-needle tasks, MQ and MV stay at or below 36\% for all models, including softmax attention, while MK-1 improves mainly under Repeat with the exact prefix. MK-2 and MK-3 are omitted from \autoref{tab:niah-multi}: the recurrent models generate almost no correct answers, whereas softmax attention reaches up to 99\% on MK-2 and 62\% on MK-3 with the exact prefix.

\begin{table}[p]
  \centering
  \caption{Single-needle NIAH accuracy (\%) per task. S, C, and P are exact-match generation accuracy with the standard, colon-appended, and exact-prefix prompts; T is teacher-forced token accuracy under the colon prompt. R (Repeat) and L (TLCE) are applied during cooldown; Full applies both throughout training. Softmax was not evaluated at 8K, and NIAH-4 T was not run for the baselines at 8K.}
  \label{tab:niah-single}
  \scriptsize
  \setlength{\tabcolsep}{3.5pt}
  \begin{tabular}{@{}llrrrrrrrrrrrrrrrr@{}}
    \toprule
    & & \multicolumn{4}{c}{NIAH-1} & \multicolumn{4}{c}{NIAH-2} & \multicolumn{4}{c}{NIAH-3} & \multicolumn{4}{c}{NIAH-4} \\
    \cmidrule(lr){3-6}\cmidrule(lr){7-10}\cmidrule(lr){11-14}\cmidrule(l){15-18}
    Model & Run & S & C & P & T & S & C & P & T & S & C & P & T & S & C & P & T \\
    \midrule
    \multicolumn{18}{@{}l}{\emph{2K context}} \\
    Softmax & Base & 100 & 100 & 100 & 100 & 100 & 100 & 100 & 100 & 69 & 74 & 77 & 99 & 89 & 95 & 99 & 99 \\
     & R & 100 & 100 & 100 & 100 & 100 & 100 & 100 & 100 & 94 & 96 & 98 & 100 & 94 & 100 & 100 & 100 \\
     & L & 100 & 100 & 100 & 100 & 100 & 100 & 100 & 100 & 88 & 88 & 86 & 100 & 89 & 100 & 100 & 100 \\
     & R+L & 100 & 100 & 100 & 100 & 100 & 100 & 100 & 100 & 97 & 96 & 99 & 100 & 95 & 100 & 100 & 100 \\
    \addlinespace[2pt]
    KDA & Base & 99 & 99 & 99 & 100 & 81 & 98 & 99 & 100 & 5 & 38 & 37 & 97 & 2 & 5 & 39 & 79 \\
     & R & 100 & 100 & 100 & 100 & 64 & 100 & 99 & 100 & 36 & 37 & 57 & 97 & 4 & 6 & 98 & 79 \\
     & L & 99 & 99 & 99 & 100 & 99 & 99 & 99 & 100 & 9 & 47 & 58 & 98 & 8 & 59 & 58 & 91 \\
     & R+L & 100 & 100 & 100 & 100 & 96 & 100 & 99 & 100 & 3 & 29 & 58 & 97 & 3 & 30 & 93 & 85 \\
     & Full & 100 & 100 & 100 & 100 & 99 & 100 & 100 & 100 & 85 & 87 & 89 & 100 & 18 & 77 & 100 & 95 \\
    \addlinespace[2pt]
    GDN & Base & 100 & 100 & 100 & 100 & 92 & 97 & 96 & 100 & 12 & 17 & 24 & 95 & 3 & 3 & 29 & 77 \\
     & R & 100 & 100 & 100 & 100 & 43 & 99 & 99 & 100 & 9 & 60 & 67 & 98 & 16 & 16 & 94 & 82 \\
     & L & 100 & 100 & 99 & 100 & 99 & 99 & 99 & 100 & 46 & 47 & 41 & 98 & 58 & 63 & 83 & 92 \\
     & R+L & 100 & 100 & 100 & 100 & 78 & 99 & 100 & 100 & 61 & 61 & 61 & 99 & 8 & 38 & 97 & 86 \\
    \addlinespace[2pt]
    DN & Base & 100 & 100 & 100 & 100 & 75 & 99 & 99 & 100 & 7 & 21 & 18 & 95 & 8 & 14 & 63 & 79 \\
     & R & 100 & 100 & 99 & 100 & 26 & 100 & 100 & 100 & 2 & 25 & 39 & 96 & 0 & 15 & 94 & 81 \\
     & L & 99 & 100 & 99 & 100 & 98 & 99 & 99 & 100 & 8 & 39 & 32 & 97 & 13 & 72 & 86 & 94 \\
     & R+L & 99 & 99 & 99 & 100 & 83 & 100 & 100 & 100 & 3 & 46 & 46 & 98 & 3 & 67 & 96 & 93 \\
    \addlinespace[2pt]
    GLA & Base & 10 & 22 & 78 & 90 & 64 & 97 & 96 & 100 & 2 & 6 & 3 & 90 & 6 & 7 & 21 & 69 \\
     & R & 18 & 95 & 94 & 99 & 40 & 98 & 99 & 100 & 23 & 25 & 20 & 96 & 29 & 32 & 84 & 83 \\
     & L & 99 & 99 & 99 & 100 & 99 & 99 & 84 & 100 & 23 & 23 & 11 & 95 & 63 & 63 & 61 & 91 \\
     & R+L & 92 & 99 & 99 & 100 & 76 & 100 & 100 & 100 & 29 & 30 & 28 & 96 & 58 & 63 & 88 & 91 \\
    \midrule
    \multicolumn{18}{@{}l}{\emph{4K context}} \\
    Softmax & Base & 85 & 89 & 93 & 98 & 98 & 99 & 98 & 100 & 19 & 25 & 57 & 96 & 55 & 64 & 88 & 92 \\
     & R & 97 & 97 & 99 & 100 & 100 & 100 & 100 & 100 & 48 & 60 & 91 & 99 & 79 & 83 & 100 & 96 \\
     & L & 97 & 97 & 98 & 100 & 97 & 100 & 100 & 100 & 64 & 66 & 76 & 99 & 74 & 97 & 96 & 99 \\
     & R+L & 99 & 99 & 99 & 100 & 99 & 100 & 100 & 100 & 87 & 91 & 98 & 100 & 96 & 99 & 100 & 100 \\
    \addlinespace[2pt]
    KDA & Base & 99 & 98 & 98 & 100 & 55 & 93 & 97 & 93 & 4 & 5 & 21 & 91 & 1 & 1 & 17 & 73 \\
     & R & 99 & 99 & 100 & 100 & 58 & 99 & 99 & 100 & 1 & 2 & 10 & 90 & 2 & 2 & 75 & 71 \\
     & L & 98 & 98 & 99 & 100 & 5 & 91 & 97 & 99 & 1 & 1 & 28 & 94 & 0 & 1 & 29 & 79 \\
     & R+L & 99 & 100 & 99 & 100 & 51 & 97 & 84 & 100 & 3 & 4 & 20 & 92 & 0 & 1 & 78 & 77 \\
     & Full & 100 & 100 & 100 & 100 & 40 & 62 & 99 & 95 & 12 & 18 & 57 & 96 & 4 & 24 & 98 & 84 \\
    \addlinespace[2pt]
    GDN & Base & 99 & 99 & 100 & 100 & 48 & 72 & 62 & 95 & 6 & 14 & 16 & 87 & 0 & 0 & 8 & 72 \\
     & R & 100 & 100 & 99 & 100 & 21 & 85 & 95 & 98 & 2 & 10 & 28 & 93 & 0 & 0 & 74 & 74 \\
     & L & 99 & 99 & 99 & 100 & 79 & 93 & 33 & 99 & 18 & 25 & 18 & 95 & 12 & 24 & 28 & 83 \\
     & R+L & 100 & 100 & 100 & 100 & 0 & 4 & 88 & 88 & 1 & 1 & 17 & 92 & 0 & 1 & 74 & 78 \\
    \addlinespace[2pt]
    DN & Base & 100 & 100 & 100 & 100 & 24 & 72 & 56 & 96 & 4 & 6 & 10 & 87 & 1 & 1 & 9 & 70 \\
     & R & 100 & 100 & 100 & 100 & 11 & 98 & 98 & 100 & 2 & 7 & 9 & 90 & 0 & 5 & 58 & 73 \\
     & L & 100 & 100 & 99 & 100 & 88 & 98 & 43 & 100 & 10 & 18 & 12 & 95 & 15 & 42 & 23 & 87 \\
     & R+L & 99 & 100 & 99 & 100 & 36 & 99 & 80 & 100 & 6 & 10 & 10 & 93 & 0 & 26 & 70 & 82 \\
    \addlinespace[2pt]
    GLA & Base & 2 & 4 & 20 & 66 & 30 & 66 & 71 & 94 & 3 & 4 & 5 & 74 & 9 & 9 & 27 & 56 \\
     & R & 2 & 29 & 69 & 76 & 7 & 85 & 89 & 98 & 3 & 3 & 4 & 86 & 12 & 16 & 51 & 68 \\
     & L & 36 & 68 & 88 & 95 & 82 & 79 & 50 & 89 & 2 & 2 & 5 & 88 & 28 & 28 & 39 & 80 \\
     & R+L & 7 & 94 & 96 & 99 & 56 & 90 & 94 & 99 & 3 & 2 & 3 & 90 & 13 & 33 & 58 & 83 \\
    \midrule
    \multicolumn{18}{@{}l}{\emph{8K context}} \\
    KDA & Base & 88 & 91 & 87 & 99 & 11 & 21 & 32 & 78 & 1 & 1 & 4 & 69 & 0 & 1 & 7 & -- \\
     & R & 96 & 97 & 97 & 100 & 19 & 48 & 73 & 91 & 0 & 0 & 2 & 70 & 1 & 1 & 19 & 43 \\
     & L & 92 & 92 & 94 & 99 & 23 & 51 & 46 & 92 & 0 & 0 & 6 & 88 & 0 & 0 & 8 & 68 \\
     & R+L & 96 & 98 & 99 & 100 & 21 & 61 & 57 & 94 & 0 & 0 & 0 & 80 & 0 & 0 & 14 & 60 \\
     & Full & 94 & 100 & 100 & 100 & 97 & 97 & 90 & 100 & 4 & 7 & 7 & 92 & 0 & 17 & 66 & 79 \\
    \addlinespace[2pt]
    GDN & Base & 99 & 100 & 99 & 100 & 10 & 15 & 17 & 48 & 4 & 3 & 4 & 51 & 0 & 0 & 5 & -- \\
     & R & 99 & 99 & 99 & 100 & 20 & 24 & 41 & 64 & 1 & 1 & 4 & 67 & 2 & 2 & 22 & 43 \\
     & L & 99 & 99 & 99 & 100 & 28 & 38 & 36 & 78 & 4 & 5 & 4 & 76 & 2 & 4 & 9 & 65 \\
     & R+L & 100 & 100 & 99 & 100 & 0 & 5 & 29 & 76 & 0 & 0 & 0 & 80 & 0 & 1 & 36 & 66 \\
    \addlinespace[2pt]
    DN & Base & 97 & 99 & 99 & 100 & 16 & 26 & 28 & 78 & 1 & 2 & 4 & 61 & 3 & 4 & 8 & -- \\
     & R & 99 & 100 & 99 & 100 & 3 & 27 & 54 & 87 & 0 & 2 & 2 & 70 & 1 & 6 & 17 & 44 \\
     & L & 98 & 98 & 98 & 100 & 23 & 61 & 28 & 94 & 0 & 4 & 3 & 82 & 5 & 10 & 14 & 67 \\
     & R+L & 96 & 99 & 99 & 100 & 4 & 68 & 45 & 96 & 0 & 0 & 0 & 81 & 0 & 3 & 29 & 60 \\
    \addlinespace[2pt]
    GLA & Base & 0 & 2 & 9 & 42 & 10 & 20 & 23 & 55 & 0 & 1 & 3 & 47 & 3 & 3 & 11 & -- \\
     & R & 0 & 2 & 22 & 27 & 1 & 35 & 38 & 80 & 1 & 1 & 2 & 64 & 4 & 5 & 14 & 35 \\
     & L & 2 & 7 & 14 & 59 & 9 & 46 & 26 & 88 & 2 & 2 & 3 & 73 & 9 & 9 & 17 & 58 \\
     & R+L & 5 & 28 & 69 & 83 & 0 & 24 & 44 & 85 & 0 & 0 & 0 & 78 & 0 & 3 & 20 & 57 \\
    \bottomrule
  \end{tabular}
\end{table}

\begin{table}[p]
  \centering
  \caption{Multi-needle NIAH accuracy (\%) per task, with columns and runs as in \autoref{tab:niah-single}. Teacher-forced accuracy is not available for MV. MK-2 and MK-3 are omitted because generation accuracy is approximately zero for every recurrent model at every length and prompt.}
  \label{tab:niah-multi}
  \scriptsize
  \setlength{\tabcolsep}{3.5pt}
  \begin{tabular}{@{}llrrrrrrrrrrr@{}}
    \toprule
    & & \multicolumn{4}{c}{MK-1} & \multicolumn{4}{c}{MQ} & \multicolumn{3}{c}{MV} \\
    \cmidrule(lr){3-6}\cmidrule(lr){7-10}\cmidrule(l){11-13}
    Model & Run & S & C & P & T & S & C & P & T & S & C & P \\
    \midrule
    \multicolumn{13}{@{}l}{\emph{2K context}} \\
    Softmax & Base & 70 & 70 & 45 & 94 & 31 & 30 & 25 & 85 & 36 & 36 & 22 \\
     & R & 52 & 53 & 81 & 94 & 29 & 29 & 25 & 85 & 26 & 27 & 24 \\
     & L & 66 & 65 & 54 & 95 & 31 & 31 & 27 & 86 & 32 & 32 & 24 \\
     & R+L & 50 & 50 & 93 & 93 & 29 & 29 & 29 & 86 & 29 & 29 & 25 \\
    \addlinespace[2pt]
    KDA & Base & 25 & 26 & 18 & 90 & 19 & 19 & 23 & 81 & 21 & 20 & 17 \\
     & R & 37 & 36 & 52 & 92 & 23 & 23 & 25 & 81 & 24 & 24 & 25 \\
     & L & 27 & 27 & 28 & 90 & 25 & 25 & 24 & 82 & 25 & 25 & 22 \\
     & R+L & 43 & 43 & 58 & 93 & 26 & 26 & 25 & 82 & 25 & 25 & 25 \\
     & Full & 59 & 59 & 72 & 95 & 28 & 28 & 26 & 84 & 27 & 27 & 25 \\
    \addlinespace[2pt]
    GDN & Base & 23 & 22 & 23 & 88 & 22 & 22 & 22 & 75 & 19 & 19 & 20 \\
     & R & 24 & 25 & 44 & 90 & 4 & 19 & 24 & 75 & 12 & 12 & 24 \\
     & L & 24 & 25 & 27 & 90 & 23 & 23 & 22 & 80 & 23 & 23 & 21 \\
     & R+L & 28 & 28 & 42 & 90 & 24 & 24 & 24 & 77 & 24 & 24 & 24 \\
    \addlinespace[2pt]
    DN & Base & 14 & 23 & 24 & 89 & 21 & 21 & 23 & 80 & 17 & 17 & 22 \\
     & R & 19 & 26 & 31 & 90 & 25 & 24 & 24 & 82 & 24 & 24 & 24 \\
     & L & 24 & 26 & 24 & 90 & 24 & 24 & 23 & 82 & 24 & 24 & 20 \\
     & R+L & 26 & 24 & 29 & 90 & 25 & 25 & 24 & 81 & 25 & 25 & 24 \\
    \addlinespace[2pt]
    GLA & Base & 20 & 22 & 19 & 87 & 16 & 17 & 20 & 75 & 18 & 18 & 16 \\
     & R & 22 & 22 & 25 & 89 & 5 & 20 & 22 & 80 & 18 & 18 & 20 \\
     & L & 21 & 21 & 19 & 88 & 20 & 20 & 20 & 79 & 20 & 20 & 18 \\
     & R+L & 16 & 16 & 21 & 88 & 21 & 22 & 22 & 81 & 23 & 23 & 21 \\
    \midrule
    \multicolumn{13}{@{}l}{\emph{4K context}} \\
    Softmax & Base & 58 & 61 & 40 & 93 & 29 & 29 & 25 & 83 & 32 & 32 & 24 \\
     & R & 52 & 54 & 91 & 94 & 26 & 26 & 25 & 84 & 27 & 27 & 24 \\
     & L & 58 & 57 & 50 & 94 & 30 & 30 & 26 & 85 & 32 & 32 & 26 \\
     & R+L & 53 & 53 & 89 & 94 & 27 & 27 & 26 & 85 & 27 & 27 & 25 \\
    \addlinespace[2pt]
    KDA & Base & 21 & 23 & 29 & 88 & 17 & 17 & 24 & 77 & 20 & 20 & 23 \\
     & R & 38 & 38 & 54 & 92 & 21 & 21 & 24 & 78 & 21 & 21 & 24 \\
     & L & 25 & 25 & 33 & 90 & 23 & 23 & 24 & 80 & 24 & 24 & 23 \\
     & R+L & 47 & 46 & 62 & 93 & 22 & 22 & 24 & 80 & 23 & 23 & 24 \\
     & Full & 51 & 52 & 64 & 93 & 27 & 29 & 26 & 81 & 27 & 27 & 25 \\
    \addlinespace[2pt]
    GDN & Base & 18 & 21 & 25 & 80 & 9 & 16 & 21 & 62 & 13 & 13 & 22 \\
     & R & 27 & 27 & 37 & 88 & 6 & 11 & 23 & 69 & 5 & 5 & 23 \\
     & L & 21 & 21 & 24 & 87 & 21 & 21 & 22 & 74 & 19 & 19 & 22 \\
     & R+L & 22 & 21 & 37 & 87 & 19 & 19 & 23 & 74 & 3 & 3 & 24 \\
    \addlinespace[2pt]
    DN & Base & 11 & 24 & 25 & 87 & 16 & 16 & 24 & 76 & 12 & 12 & 23 \\
     & R & 10 & 27 & 33 & 90 & 23 & 23 & 23 & 79 & 21 & 21 & 23 \\
     & L & 24 & 23 & 23 & 89 & 18 & 23 & 22 & 77 & 23 & 22 & 22 \\
     & R+L & 24 & 25 & 29 & 90 & 24 & 24 & 23 & 77 & 23 & 23 & 23 \\
    \addlinespace[2pt]
    GLA & Base & 17 & 20 & 17 & 82 & 9 & 9 & 14 & 63 & 16 & 16 & 14 \\
     & R & 15 & 20 & 26 & 87 & 9 & 12 & 21 & 76 & 13 & 14 & 19 \\
     & L & 17 & 17 & 18 & 86 & 18 & 18 & 16 & 72 & 18 & 18 & 14 \\
     & R+L & 20 & 20 & 23 & 87 & 17 & 18 & 18 & 77 & 20 & 20 & 19 \\
    \midrule
    \multicolumn{13}{@{}l}{\emph{8K context}} \\
    KDA & Base & 7 & 19 & 25 & 80 & 3 & 3 & 19 & 69 & 5 & 5 & 20 \\
     & R & 31 & 34 & 51 & 89 & 10 & 10 & 22 & 71 & 10 & 10 & 23 \\
     & L & 23 & 24 & 27 & 87 & 18 & 17 & 20 & 76 & 19 & 19 & 20 \\
     & R+L & 40 & 40 & 56 & 92 & 17 & 16 & 23 & 76 & 17 & 17 & 23 \\
     & Full & 48 & 49 & 55 & 93 & 22 & 22 & 24 & 79 & 24 & 24 & 24 \\
    \addlinespace[2pt]
    GDN & Base & 8 & 13 & 13 & 49 & 2 & 2 & 13 & 39 & 1 & 1 & 11 \\
     & R & 18 & 19 & 27 & 71 & 3 & 6 & 19 & 52 & 3 & 3 & 20 \\
     & L & 21 & 22 & 22 & 76 & 3 & 5 & 20 & 57 & 4 & 4 & 19 \\
     & R+L & 7 & 16 & 24 & 77 & 3 & 4 & 19 & 60 & 1 & 1 & 18 \\
    \addlinespace[2pt]
    DN & Base & 10 & 18 & 19 & 75 & 5 & 8 & 19 & 64 & 6 & 7 & 20 \\
     & R & 2 & 19 & 25 & 85 & 1 & 12 & 20 & 72 & 1 & 11 & 19 \\
     & L & 14 & 21 & 25 & 86 & 1 & 19 & 21 & 75 & 6 & 19 & 21 \\
     & R+L & 5 & 21 & 24 & 88 & 17 & 17 & 19 & 77 & 17 & 17 & 20 \\
    \addlinespace[2pt]
    GLA & Base & 10 & 15 & 15 & 58 & 3 & 3 & 14 & 45 & 9 & 9 & 14 \\
     & R & 3 & 16 & 17 & 79 & 2 & 6 & 15 & 65 & 6 & 6 & 14 \\
     & L & 14 & 18 & 17 & 81 & 15 & 15 & 16 & 63 & 16 & 16 & 16 \\
     & R+L & 6 & 10 & 15 & 82 & 8 & 8 & 14 & 71 & 14 & 14 & 15 \\
    \bottomrule
  \end{tabular}
\end{table}

\end{document}